\documentclass[journal]{IEEEtai}

\usepackage[colorlinks,urlcolor=blue,linkcolor=blue,citecolor=blue]{hyperref}

\usepackage{color,array}

\usepackage{graphicx}

\usepackage{multirow}
\usepackage{booktabs}

\usepackage[utf8]{inputenc}    
\usepackage{newunicodechar}    
\newunicodechar{，}{,}         
\newunicodechar{ℓ}{\ensuremath{\ell}}  

\usepackage{comment}
\usepackage{amsmath}
\usepackage{fancyhdr}    
\begin{document}

\title{Vision-Based Anti Unmanned Aerial Technology: Opportunities and Challenges}


\author{Guanghai Ding$^{\dagger}$, Yihua Ren$^{\dagger}$, Yuting Liu, Qijun Zhao, and Shuiwang Li$^*$\thanks{* Corresponding author. $^{\dagger}$ These authors contributed equally.}

\thanks{Guanghai Ding and Shuiwang Li are with the College of Computer Science and Engineering, Guilin University of Technology, China, 541006 and Guangxi Key Laboratory of Embedded Technology and Intelligent System, Guilin University of Technology, China, 541004 (e-mail: dingguanghai2025@163.com; lishuiwang0721@163.com).}
\thanks{Yihua Ren is with the School of Artificial Intelligence, Optics and Electronics (iOPEN), Northwestern Polytechnical University, Xi'an 710072, China (e-mail: renyihua@mail.nwpu.edu.cn).} 
\thanks{Yuting Liu is with JD Logistic, Beijing, China (e-mail: liuyuting67@jd.com).} 
\thanks{Qijun Zhao is with the College of Computer Science, Sichuan University, Sichuan 610065, China (e-mail: qjzhao@scu.edu.cn).}

}


\markboth{}
{}

\maketitle
\begin{abstract}
With the rapid advancement of UAV technology and its extensive application in various fields such as military reconnaissance, environmental monitoring, and logistics, achieving efficient and accurate Anti-UAV tracking has become essential. The importance of Anti-UAV tracking is increasingly prominent, especially in scenarios such as public safety, border patrol, search and rescue, and agricultural monitoring, where operations in complex environments can provide enhanced security. Current mainstream Anti-UAV tracking technologies are primarily centered around computer vision techniques, particularly those that integrate multi-sensor data fusion with advanced detection and tracking algorithms.
This paper first reviews the characteristics and current challenges of Anti-UAV detection and tracking technologies. Next, it investigates and compiles several publicly available datasets, providing accessible links to support researchers in efficiently addressing related challenges. Furthermore, the paper analyzes the major vision-based and vision-fusion-based Anti-UAV detection and tracking algorithms proposed in recent years. Finally, based on the above research, this paper outlines future research directions, aiming to provide valuable insights for advancing the field.
\end{abstract}


\begin{IEEEkeywords}
Anti-UAV, Anti-UAV Datasets, Anti-UAV Methods, Computer Vision
\end{IEEEkeywords}

\section{Introduction}

\IEEEPARstart{I}{n} recent years, the Unmanned Aerial Vehicle (UAV) industry has undergone exponential growth, establishing itself as a pivotal catalyst for global economic development and technological advancement \cite{tepylo2023public}. Projections indicate that by 2025, this sector is poised to emerge as a trillion-dollar market ecosystem, with particularly accelerated expansion observed in low-altitude applications, thereby cementing its position as a cornerstone of the new economic paradigm. Concurrently with these technological breakthroughs and mass-market penetration, associated security vulnerabilities have escalated disproportionately, necessitating urgent advancements in  Anti-UAV technologies \cite{mei2025unsupervised}.

\begin{figure*}[h]
\centering
\includegraphics[width=1\linewidth]{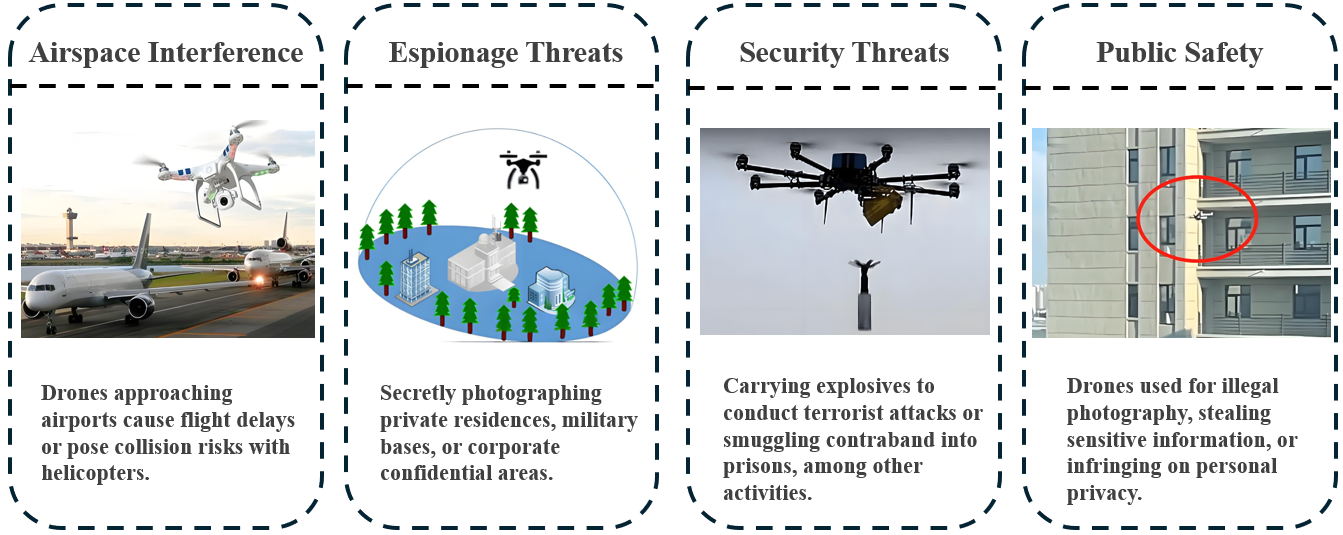}
\caption{Major categories of UAV threats:Airspace Interference, Espionage Threats, Security Threats, Public Safety.}
\label{newFig1}
\end{figure*}

The ubiquitous proliferation of Unmanned Aerial Vehicle (UAV) technologies has catalyzed their diversified deployment across dual-use (civilian-military) operational domains. In civilian applications, UAV systems demonstrate transformative potential through three principal implementations: (1) emergency response coordination in disaster scenarios \cite{wang2023secure}, (2) precision agricultural monitoring employing multispectral imaging \cite{khaki2022wheatnet}, and (3) intelligent traffic management via aerial surveillance networks \cite{krichen2022security}, collectively enhancing operational efficiency through streamlined data acquisition processes   \cite{liu2024vision}. 
Notably, this technological dual-use nature has precipitated substantial security externalities. More alarmingly, modern UAV platforms have been weaponized for nefarious purposes, including but not limited to: (a) conducting illicit surveillance operations that violate personal privacy through advanced computer vision techniques  \cite{al2024deep}, and (b) facilitating contraband transportation via autonomous navigation systems, particularly exacerbating narcotics trafficking and weapons proliferation risks \cite{lee2023machine}.
From a military-strategic perspective, UAV fleets are being increasingly integrated into network-centric warfare architectures for multi-role combat operations encompassing ISR (Intelligence, Surveillance, Reconnaissance), target acquisition, and precision strike missions \cite{wei2022perdet}. The convergence of stealth-capable UAV designs with complex electromagnetic terrain, as illustrated in Figure \ref{newFig1}, creates asymmetric warfare complexities that fundamentally challenge conventional air defense paradigms. These developments underscore the critical imperative for advancing Detection and Tracking Systems for UAV to ensure comprehensive security assurance across both civil infrastructure and national defense frameworks.

Recently, numerous emerging research fields have successively emerged. For instance, in \cite{cai2025biomimetic}, researchers conducted an in-depth investigation into the collective coordination capabilities of swarms through biologically inspired swarm-splitting algorithms. By employing a small number of "driver agents," the target UAV swarm can be forcibly segmented into multiple sub-swarms, thereby disrupting its collaborative capacity to achieve countermeasure effects. In \cite{wei2024survey}, blockchain technology was integrated with anti-UAV strategies, where data synchronization within UAV swarms was interfered with via blockchain to realize countermeasures. Additionally, \cite{wang2024thinking} highlighted that quantum radar, leveraging quantum entanglement detection and micro-Doppler analysis, can address the detection blind spots of traditional radar systems in complex electromagnetic environments. This advancement enables quantum radar to serve as a more effective early-warning system for detecting UAV targets.

Anti-UAV detection is the primary step in identifying potential threats, with its core focus on the rapid discovery and confirmation of the presence of UAV in the airspace \cite{svanstrom2022drone}. The detection system integrates multi-source data from radar, radio frequency signal analysis, and electro-optical sensors, enabling it to distinguish UAV from interference targets such as birds and kites in complex environments, while providing real-time alerts \cite{aposporis2020object,li2022radar,sun2024deep}. Anti-UAV tracking is a key technology for addressing dynamic threats, aiming to continuously lock onto the target’s location and predict its movement trajectory \cite{huang2023anti}. UAV often possess high-speed maneuverability, low-altitude flight, and active evasion capabilities, making it challenging for traditional monitoring methods to maintain continuous capture \cite{yuan2023thermal,marvasti2021deep}. Tracking algorithms utilize multi-sensor data fusion to calculate the target’s three-dimensional coordinates, speed, and heading in real-time, especially maintaining stable tracking during brief occlusions or changes in the target’s shape \cite{gao2023drone,kumar2024correlation}.

To address the threats depicted in Figure \ref{newFig1}, researchers have proposed various technological approaches \cite{li2024robust}, including radar detection, radio signal analysis, acoustic sensing, and computer vision, among others. Radar detection technology \cite{coluccia2020detection,kumawat2022diat,wang2021deep} is not affected by lighting conditions and can operate in all weather, but its ability to distinguish low-altitude, slow-moving small targets and UAV is limited. Electro-optical imaging technology \cite{wang2022survey,zhai2023yolo,xu2024anti} captures the thermal radiation signals of targets through infrared cameras or thermal imaging devices, but its resolution may decrease in low light conditions. Radio signal \cite{sharma2024advanced,alam2023rf,jurn2024review} analysis technology locates and tracks UAV by capturing their communication signals, but it is ineffective against UAV in stealth mode. Acoustic sensing technology \cite{utebayevapractical,dong2023drone,kummritz2024sound} detects UAV by analyzing the noise characteristics they produce, but it is significantly affected by environmental noise and has limited detection range. In recent years, the rapid development of artificial intelligence technologies has provided new solutions for Anti-UAV tracking. Deep learning-based object detection and tracking algorithms \cite{wei2023review,huang2023anti} can extract target features from complex backgrounds, improving detection accuracy and real-time performance. The integration of multimodal sensor fusion technologies \cite{wei2025uav,chauhan2025nation}, including infrared imaging, thermal imaging, radar, and visible light cameras, with deep learning-based object detection and tracking algorithms, presents a new opportunity for Anti-UAV detection and tracking.


This paper aims to summarize the latest advancements in the field of Anti-UAV technologies to help researchers and engineers better understand and apply these techniques, thereby developing more efficient and targeted models. The main contributions of this paper include the following aspects:

\begin{itemize}  
  \item This paper analyzes the key characteristics and existing challenges of Anti-UAV technologies, while also highlighting the main features and challenges of Anti-UAV detection and tracking. 
  \item This paper compiles high-quality RGB and IR datasets for visual-based Anti-UAV research and provides effective links to publicly available datasets to facilitate efficient development for researchers.
  \item We summarize and analyze the latest vision-based and vision-fusion-based anti-UAV detection and tracking methods, analyze the limitations of the methods and datasets, and propose seven potential research directions for future anti-UAV technologies.
\end{itemize}

\section{Main features of the dataset and method.}
Anti-UAV detection and tracking have evolved from radar, optical, acoustic, and radio frequency (RF) detection techniques to computer vision-based detection and tracking methods integrated with deep learning \cite{liu2024vision,zhao2021unified,ince2021semi}. This section focuses on describing the main features of the datasets and methods for Anti-UAV detection and tracking based on vision and visual fusion combined with deep learning.

\subsection{Main features of the dataset}

Since Anti-UAV detection and tracking based on computer vision collects data using visible cameras and infrared (IR) sensors \cite{koksal2020effect,liu2020ipg,wu2020real}, current datasets are primarily categorized into RGB datasets and IR datasets \cite{pawelczyk2020real}. Under well-illuminated conditions, visible light imagery typically provides substantial discriminative features (e.g., chromatic attributes, textural properties, and surface pattern characteristics). In low-light or nighttime scenarios, infrared sensors capture images based on the thermal radiation contrast between the target and the background, enabling effective operation in low-light or nighttime conditions, albeit with lower image contrast and resolution \cite{zhang2025novel}.

\begin{figure}[h]
\centering
\includegraphics[width=1\linewidth]{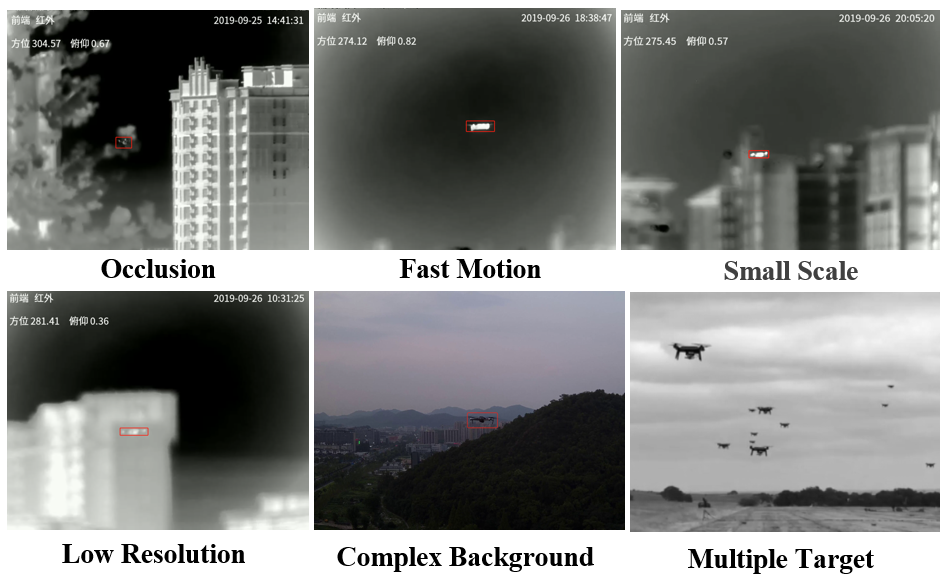}
\caption{Some characteristics of UAV detection and tracking, with the Chinese characters in the image indicating the shooting time and location.}
\label{newFig2}
\end{figure}

Compared with detecting and tracking other objects, Anti-UAV detection and tracking exhibit some unique characteristics, as shown in Figure \ref{newFig2}. First, most consumer UAV are typically small in size, fly at low altitudes, and possess inconspicuous target features, making them prone to misclassification, especially in complex backgrounds \cite{isaac2021unmanned,fang2021real}. Second, the rapid movement and sudden direction changes of UAV during flight result in significant variations in target size and shape, increasing the likelihood of target loss \cite{fang2020infrared}. Additionally, since UAV operate outdoors, they are susceptible to interference from dynamic outdoor backgrounds and environmental noise \cite{jiang2021anti}, particularly in thermal infrared scenarios where discriminative features are often lacking.

\subsection{Main features of the method}

Building upon these dataset characteristics, current Anti-UAV detection and tracking methods have developed distinct technical features to address these challenges. The primary characteristics are reflected in multi-technology fusion, cross-modal collaboration, and lightweight adaptation \cite{koulouris2023comparative,chen2022benchmark}.

On one hand, traditional sensor-based approaches rely on multi-source heterogeneous data fusion to achieve robust perception in non-line-of-sight environments. However, these methods come with high hardware deployment costs and limited anti-interference capabilities. On the other hand, deep learning-based vision methods \cite{chen2025strong,du2025f} significantly improve detection accuracy and real-time performance in complex backgrounds through small-object feature enhancement, dynamic template matching, and global attention mechanisms \cite{yu2025yolo}.

Notably, cross-modal collaboration techniques \cite{jouaber2021nnakf} have emerged as a promising direction, enhancing perception by aligning and complementing features across different modalities. While these methods show superior performance, they require careful balancing between computational cost and real-time processing constraints.

Based on a comprehensive investigation and analysis of the latest literature, this paper focuses on vision-based methods and categorizes them into two groups according to their template matching mechanisms, real-time detection tasks, lightweight network structures, global dependency modeling via self-attention mechanisms, and cross-modal feature complementarity. These categories include UAV-based Detection methods and UAV-based Tracking methods.

\section{Anti-UAV Datasets}
To address the challenges associated with Anti-UAV tasks, researchers have constructed datasets based on visual, infrared, radio frequency, and radar data. To assist researchers in training large models, this section primarily discusses open-source computer vision-based datasets in chronological order.

\textbf{The 4th Anti-UAV Competition \cite{4thAntiUAV}} continued to build upon the benchmark dataset established in previous editions. This dataset consists of 410 high-quality video sequences and has been further expanded by introducing new challenge tasks, such as multi-UAV tracking, thereby broadening the scope of Anti-UAV tasks. The competition included three tracks: Track 1: Single UAV Tracking – Given the UAV's bounding box in the initial frame, the algorithm must track the target and predict its bounding box, marking it as invisible when the target disappears. Track 2: Single UAV Detection and Tracking – The target is unknown in the initial frame; the algorithm must detect and track the UAV, predict its bounding box, and mark it as invisible when the target is absent or disappears. Track 3: Multi-UAV Tracking – Given the bounding boxes of multiple UAV in the initial frame, the algorithm must track both the initial and newly appearing targets, predict their bounding boxes, and assign unique IDs.

\textbf{The Anti2 dataset \cite{gao2024novel}} introduces multiple types of aerial objects, including UAV, airplanes, helicopters, and birds, comprising a total of 5,062 images. It provides a more challenging test platform for evaluating the Anti-UAV detector’s interference resistance capabilities. The dataset sources images from diverse origins, including: Open-source object detection datasets,Open-source object tracking datasets, Video platforms such as Bilibili and YouTube, Text-based search engines like Google, Bing, and Baidu, as well as image search engines such as Yandex and TinEye. The dataset is annotated using the LabelImg tool, with a total of 5,828 bounding boxes labeled, including 1,688 UAV, 1,444 birds, 1,406 helicopters, and 1,320 airplanes. The annotation information is stored in YOLO format, making it compatible with all YOLO-series algorithms. Most objects in this dataset are small-scale flying targets, closely replicating real-world Anti-UAV application scenarios.

\textbf{The MMFW-UAV dataset \cite{liu2025mmfw}} is a large-scale multi-modal benchmark tailored for visual tasks involving fixed-wing UAV. It contains 147,417 high-quality images, evenly divided into zoomed optical, wide-angle optical, and thermal infrared modalities. Data were collected using a DJI M30T UAV during real-world flights, covering 12 fixed-wing UAV types with varying sizes, structures, and appearances. Images were captured from top-down, side, and bottom-up angles under diverse lighting and weather conditions. Annotations follow Pascal VOC and MS COCO formats, created through semi-automatic labeling, manual verification, and refinement. The dataset offers temporally and spatially aligned multi-modal image pairs, supporting tasks like object detection, tracking, and multi-modal fusion.

\begin{figure*}[h]
\centering
\includegraphics[width=1\linewidth]{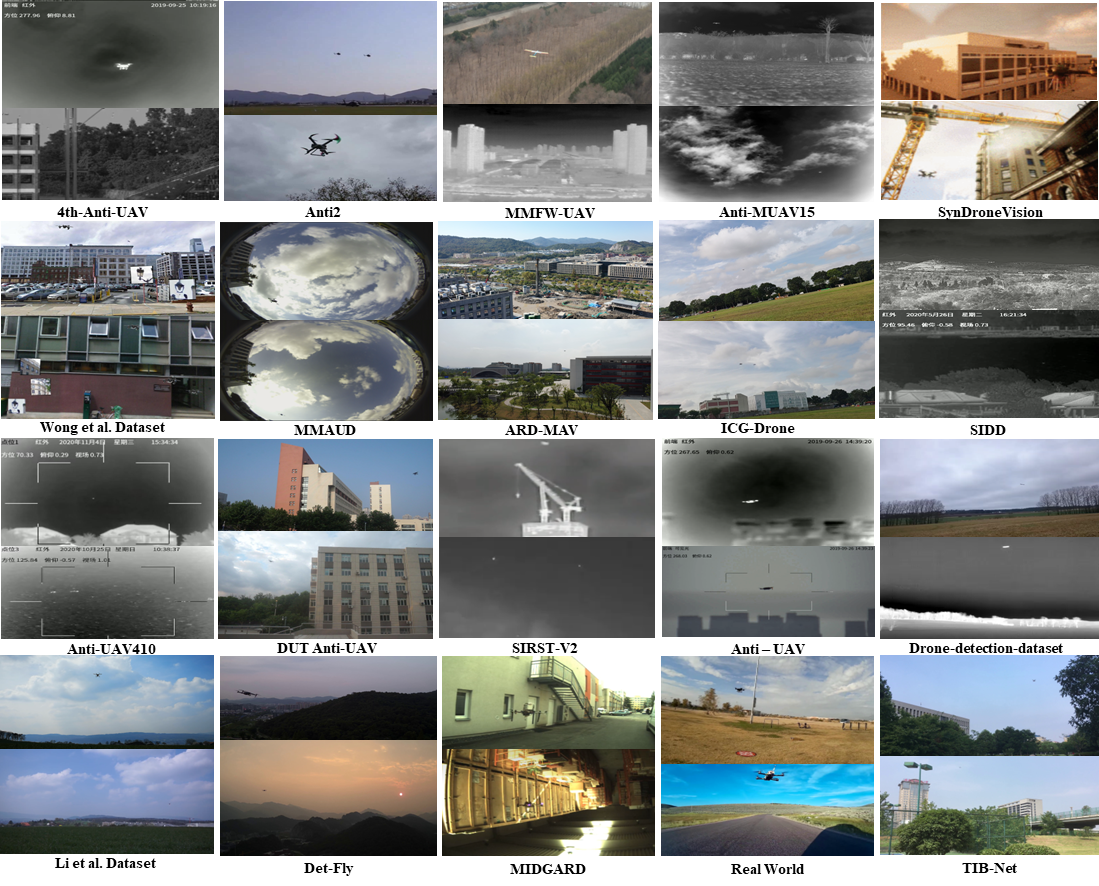}
\caption{Some example images from the aforementioned datasets.
The Chinese characters in the image represent the time and location of the shooting of the 4th Anti-UAV, SIDD, Anti-UAV410, and Anti-UAV datasets.}
\label{newFig3}
\end{figure*}

\textbf{The Anti-MUAV15 dataset \cite{liu2025learning}} is a benchmark for multi-UAV infrared tracking, containing 15 thermal video sequences with 16,269 frames captured in real-world security settings such as buildings, skies, lakes, and vegetation. Each frame includes two to three very small UAV targets, averaging 23 pixels in size, with the smallest being 12 by 5 pixels. Videos were recorded at 640×512 resolution and 15 FPS using a thermal camera. Designed for evaluating anti-UAV tracking under low visibility and multiple target conditions, it presents a realistic and challenging testbed for tracking small, fast-moving UAV in cluttered environments.

\textbf{The SynDroneVision dataset \cite{lenhard2025syndronevision}} is created using Unreal Engine 5.0 and the Colosseum platform, containing 140,038 high-resolution images (2560×1489 pixels) across eight virtual environments and thirteen UAV models, including DJI Phantom. It simulates diverse lighting conditions (e.g., day, night, sunny and cloudy) using the Lumen system and applies random blur for added variability. All images are automatically annotated with pixel-level bounding boxes, supporting end-to-end UAV detection training. The dataset provides a realistic and cost-effective resource for small object detection in complex scenes.

\textbf{The Wong et al. dataset \cite{wong2025wrn}} is a synthetic dataset containing 1,000 images. It was constructed using 100 background scenes selected from the Paris Street View dataset. On each background, three UAV of varying sizes and three non-UAV distractor objects were manually inserted to simulate realistic background interference. The dataset size was intentionally kept small to reduce overfitting and enhance generalization, and it serves as a valuable benchmark for assessing UAV detection in visually complex urban environments.

\textbf{The MMAUD dataset \cite{yuan2024mmaud}} is a multimodal Anti-UAV dataset focused on the detection, classification, and trajectory estimation of small UAV. It integrates multiple sensor inputs, including stereo vision, various radar types, and audio arrays. The dataset is available in two formats—rosbag format and file system format—to accommodate different research needs.
Notably, the stereo vision data in the MMAUD dataset is collected using stereo fisheye cameras, which offer cost-effectiveness and a wide field of view, making them highly suitable for continuous surveillance tasks.

\begin{table*}
\centering
\renewcommand{\arraystretch}{1.8}
\caption{Detailed information on the Main Scene, Type and Total, Modal, Complexity, UAV Type, and Multimodal of the above dataset}
\label{table1} 
\setlength{\tabcolsep}{-2pt}{
\resizebox{\textwidth}{!}{
\begin{tabular}{ccccccl}
\hline
\textbf{Dataset}      & \textbf{Main Scene}                                                                                   & \textbf{~~~~ Type and Total~~~ }                                                 & \textbf{~~~~~~~ Modality~~~~~~~~ } & \textbf{~~~~ Complexity ~ ~ ~ } & \textbf{UAV Type}                                                                                             & \textbf{~ ~Multimodal ~~ ~~~~~ }  \\
\hline
4th-Anti-UAV~\cite{4thAntiUAV}          & Sky,Urban,Mountain,Forest                                                                             & Video 410                                                                        & IR                                 & Large                          & Not available                                                                                                 & ~~~~~~~~~ No                          \\
\hline
Anti2~\cite{gao2024novel}                 & Sky,Urban,Mountain,Forest                                                                             & Image 5062                                                                       & RGB                                & Medium                         & Not available                                                                                                 & ~~~~~~~~~ No                          \\
\hline
MMFW-UAV~\cite{liu2025mmfw}                 & Sky,Urban,Forest                                                                             & Image 147,417                                                                       & RGB and IR                               & Large                         &  DJI M30T, Fixed-wing UAV                                                                                                 & ~~~~~~~~~ Yes                          \\
\hline
Anti-MUAV15 ~\cite{liu2025learning}                 & City, sky, lake, bushest                                                                             & video 15                                                                       & IR                                & Medium                         & Not available                                                                                                 & ~~~~~~~~~ No                          \\
\hline
SynDroneVision~\cite{lenhard2025syndronevision}                 & \begin{tabular}[c]{@{}c@{}}Campus, City, Farm,\\Industrial area \end{tabular}                                                                              & Image 140,038                                                                       & RGB                                & Large                        & \begin{tabular}[c]{@{}c@{}}DJI Phantom, DJI Tello Ryze\\and other UAVs\end{tabular}                                                                                                  & ~~~~~~~~~ No                          \\
\hline
Wong et al. dataset~\cite{wong2025wrn}                 & Crowd, Buildings                                                                             & Image 1000                                                                       & RGB                                & Small                        & Not available                                                                                                 & ~~~~~~~~~ No                          \\
\hline
MMAUD~\cite{yuan2024mmaud}                 & \multicolumn{1}{l}{\begin{tabular}[c]{@{}l@{}}Sky,Industrial,Parking,Roof,\\Urban,Noise\end{tabular}} & Multi-category~ ~~                                                               & RGB                                & Large~                         & \begin{tabular}[c]{@{}c@{}}DJl Phantom 4, DJI Avata FPV, \\DJl Mavic 3, DJL M300 and DJl Mavic 2\end{tabular} & ~~~~~~~~~ Yes                         \\
\hline
ARD-MAV~\cite{guo2024global}               & Sky, Urban,  Rural, Mountain                                                                          & \begin{tabular}[c]{@{}c@{}}Image 106, 665\\Video 60\end{tabular}                 & RGB                                & Large                          & DJI Mavic2 Pro                                                                                                & ~~~~~~~~~ Yes                         \\
\hline
ICG-Drone~\cite{xiao2024real}             & Urban, Indoor                                                                                         & \begin{tabular}[c]{@{}c@{}}Image 1, 823\\Video 5\end{tabular}                    & RGB                                & Medium                         & DJI F450, DJI Tello and QAV250                                                                                & ~~~~~~~~~ Yes                         \\
\hline
SIDD~\cite{yuan2023irsdd}                   & City, Mountain, Sea, Sky                                                                              & Image 4737                                                                       & IR                                 & Medium                         & Not available                                                                                                 & ~~~~~~~~~ No                          \\
\hline
Anti-UAV410~\cite{huang2023anti}           & Mountain,Building,Urban                                                                               & Video 410                                                                        & IR                                 & Large                          &  Not available                       & ~~~~~~~~~ No                          \\
\hline
DUT Anti-UAV~\cite{zhao2022vision}          & Sky, Jungles, Urban, Rural                                                                            & \begin{tabular}[c]{@{}c@{}}Image 1, 000\\Video 20\end{tabular}                   & RGB                                & Medium                         & Not available                                                                                                 & ~~~~~~~~ Yes                          \\
\hline
SIRST-V2~\cite{dai2023one}              & Sky, City, Sea                                                                                        & Image 1, 024                                                                     & IR                                 & Small                          & Not available                                                                                                 & ~~~~~~~~ No                           \\
\hline
Anti – UAV~\cite{jiang2021anti}            & Sky,Urban,Mountain                                                                                    & \begin{tabular}[c]{@{}c@{}}Video\_RGB 318 \\Video\_IR 318\end{tabular}           & RGB and IR                         & Large                          & DJI and Parrot                                                                                                & ~~~~~~~~ Yes                          \\
\hline
UAV-Detection-Dataset~\cite{svanstrom2021dataset} & Airport                                                                                               & \begin{tabular}[c]{@{}c@{}}Audio 90\\Video\_RGB 285 \\Video\_IR 365\end{tabular} & RGB and IR                         & Large                          & \begin{tabular}[c]{@{}c@{}}DJI Phantom4 Pro,DJI Flame Wheel \\ and Hubsan H107D+\end{tabular}                 & ~~~~~~~~ Yes                          \\
\hline
Li et al. Dataset~\cite{li2020reconstruction}     & Sky,Lawn,Forest                                                                                       & Video 27                                                                         & RGB                                & Medium                         & Yuneec                                                                                                        & ~~~~~~~~ No                           \\
\hline
Det-Fly~\cite{zheng2021air}               & Sky, Field, Urban, Mountain                                                                           & Image 13, 271                                                                    & RGB                                & ~ ~ Medium~ ~~                 & DJI M210 and DJI Mavic                                                                                        & ~~~~~~~~ No                           \\
\hline
MIDGARD~\cite{walter2020training}                 & Fields, Urban, Indoor,Forest                                                                          & Image 7, 595                                                                     & RGB                                & Medium                         & Not available                                                                                                 & ~~~~~~~~ No                           \\
\hline
Real World~\cite{pawelczyk2020real}            & Sky, Urban,  Rural                                                                                    & Image 56, 821                                                                    & RGB                                & Medium                         & Not available                                                                                                 & ~~~~~~~~ No                           \\
\hline
TIB-Net~\cite{sun2020tib}               & College, illumination                                                                                 & Image 2, 860                                                                     & RGB                                & Small                          & Not available                                                                                                 & ~~~~~~~~ No                           \\
\hline
\end{tabular}

}}
\end{table*}

\textbf{The ARD-MAV dataset \cite{guo2024global}} consists of 60 video sequences and 106,665 image frames. The videos were captured using DJI Mavic 2 Pro and M300 cameras during low-altitude and mid-altitude flights, covering real-world challenges such as complex backgrounds, non-planar scenes, occlusions, sudden camera movements, rapid motion, and small MAVs.
Each video lasts approximately 1 minute, with a frame rate of 30 FPS and a resolution of 1920×1080, containing only a single MAV. All targets were manually annotated using the labelImg tool, with target sizes ranging from 6×3 to 136×75 pixels. The smallest target was an MAV captured from 150 meters away, with an average size occupying only 0.02\% of the image—making it the dataset with the smallest average MAV target size among existing MAV datasets.

\textbf{The ICG-Drone  dataset \cite{xiao2024real}} consists of two UAV datasets: ICG-Drone-2c, which includes 1,575 images of two UAV types (Tello and F450), and ICG-Drone-3c, which contains 1,823 images of three UAV types (Tello, F450, and QAV).
The dataset was constructed by integrating original video recordings and internet-sourced videos featuring DJI Tello, DJI F450, and QAV250 UAV. All images are standardized to a resolution of 1920 × 1080 pixels. The dataset is split into training and validation sets at an 8:2 ratio, with the ICG-Drone-3c training set containing 1,457 images and the validation set containing 366 images.
This dataset supports multi-class UAV detection in both indoor and outdoor environments and includes video data for tracking tasks, making it a valuable resource for multi-class Anti-UAV tracking in diverse scenarios.

\begin{table*}
\centering
\renewcommand{\arraystretch}{1.5}
\caption{Detailed information on the source, public links, and access time of the dataset}
\label{table2} 
\setlength{\tabcolsep}{-2pt}{
\resizebox{\textwidth}{!}{
\begin{tabular}{ccccc}
\toprule
\hline
\textbf{Dataset}      & \textbf{~~~~~ Year~ } & \textbf{~~~~~~~~~~~ Source~~~~~~~~~~ } & \textbf{Public Link}                                                 & \textbf{~~ ~~~~ Access Data~ ~ ~~ ~ }  \\
\hline
4th-Anti-UAV~\cite{4thAntiUAV}          & ~~~~~2025                  & CVPR                                   & https://anti-uav.github.io/                                          & 26-Feb-25                                  \\
\hline
Anti2~\cite{gao2024novel}                 & ~~~~~2025                  & MDPI                                   & https://github.com/gdpinntit/-anti-interference-and-anti-UAV-dataset & 26-Feb-25                                  \\
\hline
MMFW-UAV~\cite{liu2025mmfw}                 & ~~~~~2025                  & SCI DATA                                   & https://doi.org/10.57760/sciencedb.07839 & 19-Jun-25                                  \\
\hline
Anti-MUAV15~\cite{liu2025learning}                 & ~~~~~2025                  & ~~EXPERT SYST APPL                                  & https://github.com/Shihan0325/Anti-MUAV15 & 19-Jun-25                                  \\
\hline
SynDroneVision~\cite{lenhard2025syndronevision}                 & ~~~~~2025                  & WACV                                   & https://zenodo.org/records/13360116 & 18-Jun-25                                  \\
\hline
Wong et al. dataset~\cite{wong2025wrn}                 & ~~~~~2025                  & IJCNN                                   & https://github.com/yjwong1999/IJCNN2025-DvB & 18-Jun-25                                  \\
\hline
MMAUD~\cite{yuan2024mmaud}                 & ~~~~~2024                  & ICRA                                   & https://github.com/ntu-aris/MMAUD                                    & 6-Jan-25                                   \\
\hline
ARD-MAV~\cite{guo2024global}               & ~~~~~2024                  & IEEE TITS                              & https://github.com/WindyLab/Global-Local-MAV-Detection               & 6-Jan-25                                   \\
\hline
ICG-Dron~\cite{xiao2024real}              & ~~~~~2024                  & IEEE T-IV                              & https://github.com/NTU-ICG/multiUAV -detection-tracking              & 8-Nov-24                                   \\
\hline
SIDD~\cite{yuan2023irsdd}                  & ~~~~~2023                  & MDPI                                   & https://github.com/Dang-zy/SIDD                                      & 8-Nov-24                                   \\
\hline
Anti-UAV410~\cite{huang2023anti}           & ~~~~~2023                  & IEEE TPAMI                             & https://github.com/HwangBo94/Anti-UAV410                             & 8-Nov-24                                   \\
\hline
DUT Anti-UAV~\cite{zhao2022vision}          & ~~~~~2022                  & IEEE TITS                              & https://github.com/wangdongdut/DUT-Anti-UAV                          & 8-Nov-24                                   \\
\hline
SIRST-V2~\cite{dai2023one}              & ~~~~~2022                  & IEEE TGRS                              & https://github.com/YimianDai/open-sirst-v2                           & 8-Nov-24                                   \\
\hline
Anti–UAV~\cite{jiang2021anti}              & ~~~~~2021                  & IEEE TMM                               & https://github.com/ucas-vg/Anti-UAV                                  & 24-Sep-24                                  \\
\hline
UAV-Detection-Dataset~\cite{svanstrom2021dataset} & ~~~~~2021                  & Data Brief                             & https://github.com/UAV DetectionThesis/UAV -detection-dataset        & 6-Dec-24                                   \\
\hline
Li et al. Dataset~\cite{li2020reconstruction}     & ~~~~~2020                  & IROS                                   & https://github.com/CenekAlbl/UAV -tracking-datasets                  & 24-Sep-24                                  \\
\hline
Det-Fly~\cite{zheng2021air}               & ~~~~~2020                  & IEEE RA-L                              & https://github.com/Jake-WU/Det-Fly                                   & 6-Dec-24                                   \\
\hline
MIDGARD~\cite{walter2020training}               & ~~~~~2020                  & ICRA                                   & https://mrs.felk.cvut.cz/midgard                                     & 6-Dec-24                                   \\
\hline
Real World~\cite{pawelczyk2020real}            & ~~~~~2020                  & IEEE Access                            & https://github.com/Maciullo/UAV DetectionDataset/tree/master         & 6-Dec-24                                   \\
\hline
TIB-Net~\cite{sun2020tib}               & ~~~~~2020                  & IEEE Access                            & https://github.com/kyn0v/TIB-Net                                     & 6-Dec-24                                   \\
\bottomrule
\end{tabular}

}}

\end{table*}

\textbf{The SIDD dataset \cite{yuan2023irsdd}} focuses on the detection of quadcopter UAV in infrared imaging, which typically exhibit the characteristics of being low-altitude, slow-speed, and small-sized ("low, slow, and small"). It comprises 4,737 infrared images with a resolution of 640 × 512 pixels, systematically constructing four representative scenarios: mountainous terrain, urban environments, sky, and ocean to address the challenges of UAV detection in complex backgrounds.
Due to the nature of infrared imaging, UAV targets often have blurred edge information, and each environment introduces unique interference sources—such as terrain heat radiation in mountains and building heat sources in urban areas—closely resembling real-world UAV intrusion scenarios.
The dataset is split into an 8:2 ratio for training and testing, providing a standard benchmark for evaluating algorithm generalization across diverse environments. With its multi-scenario coverage and precise annotation, SIDD serves as a valuable dataset for studying the challenges of detecting "low, slow, and small" infrared targets in complex backgrounds.

\textbf{The Anti-UAV410 Dataset \cite{huang2023anti}} is a mid-wave infrared (TIR) video dataset designed to simulate real-world Anti-UAV tracking challenges. It encompasses a wide variety of complex scenes, including two lighting conditions (day and night), two seasons (autumn and winter), and six types of background environments such as buildings (30\%), mountains (20\%), and urban areas (30\%), all captured at a frame rate of 25 FPS.
The dataset contains 150,000 frames of meticulously annotated core video, totaling 100 minutes. It has been expanded by integrating the first Anti-UAV Challenge \cite{4thAntiUAV} and the Anti-UAV multimodal dataset \cite{jiang2021anti}, and has been refined and cleaned to form the final version. The average length of the videos is 1,069 frames, focusing on testing the long-term target re-detection capabilities of trackers.
This dataset breaks away from traditional partitioning methods by categorizing the 410 sequences based on scene attributes into 200 training sets, 90 validation sets, and 120 test sets, emphasizing the evaluation of generalization in practical applications. Notably, the dataset features a target occlusion rate exceeding 60\% and a scene proportion of sudden lighting changes at 35\%, significantly enhancing the credibility of evaluations for UAV tracking tasks in complex environments.

\textbf{The DUT Anti-UAV dataset \cite{zhao2022vision}} is a Anti-UAV dataset that includes both detection and tracking subsets. he detection subset consists of 10,000 images, which are divided into a training set (5,200 images), validation set (2,600 images), and test set (2,200 images). The image resolution ranges from a maximum of 3744 × 5616 pixels to a minimum of 160 × 240 pixels. The tracking subset contains 20 video sequences, covering both short-term and long-term tracking scenarios. The video resolutions are 1080 × 1920 and 720 × 1280 pixels. This dataset is well-suited for both short-term and long-term Anti-UAV tracking tasks, providing a diverse and comprehensive benchmark for Anti-UAV research.

\textbf{The SIRST-V2 dataset \cite{dai2023one}} is an updated version of the previous \textbf{SIRST-V1 dataset \cite{dai2021asymmetric}}, containing 1,024 images, most of which have a resolution of 1280 × 1024 pixels. These images are extracted from real-world scene videos, providing a challenging benchmark for infrared small target detection in complex environments.
A notable enhancement in SIRST-V2 is the inclusion of urban scenes, which introduce a significant amount of background clutter resembling the target objects. Traditional methods based on target saliency or low-rank sparse decomposition struggle to effectively identify targets in such conditions. Instead, deep networks with high-level semantic understanding are required to accurately distinguish targets from non-target interference.

\textbf{The Anti-UAV  dataset \cite{jiang2021anti}} introduces a new perspective on Anti-UAV tracking, comprising 318 paired videos with over 580k manually annotated bounding boxes. Each video pair consists of one RGB video and one infrared video, with the primary target being UAV, most of which are large-sized—a contrast to the small UAV commonly encountered in real-world scenarios. The training and validation sets are derived from non-overlapping segments of the same videos, whereas the test set consists of entirely different videos, featuring higher complexity. Among the 318 video pairs, 160 are allocated for training, 91 for testing, and the remaining for validation. The dataset primarily covers tracking scenarios in sky and urban environments, but lacks data representing complex natural environments or extreme weather conditions.

\textbf{The UAV-Detection-Dataset \cite{svanstrom2021dataset}} is a large-scale dataset captured at three airports in Sweden, utilizing three different UAV models for video recording. It comprises 90 audio recordings, 285 RGB videos, and 365 infrared videos, making it a comprehensive resource for UAV detection and tracking tasks. The dataset incorporates infrared technology, making it particularly suitable for nighttime UAV detection and tracking applications. The dataset features multiple UAV types, including DJI Phantom 4 Pro, DJI Flame Wheel, and Hubsan H107D+, ensuring diverse UAV characteristics for robust model evaluation.

\textbf{The Li et al. Dataset \cite{li2020reconstruction}} utilizes a Yuneec hexacopter UAV equipped with a Fixposition RTK GNSS system, offering a positioning accuracy of ±1 cm. The UAV flies within a 100×100 meter area, reaching heights of up to 50 meters, with 4-7 cameras (smartphones, compact cameras, GoPro) set up around the area to ensure the UAV is visible from at least two angles. The dataset consists of 5 sub-datasets divided into 27 long video sequences. Datasets 1 and 2: The UAVflies at a moderate speed, with smooth direction changes. Dataset 3: Presents greater challenges, with increased speed, acceleration, and longer flight paths, covering various flight modes such as straight flight, sharp turns, and rapid ascents and descents. Dataset 4: The most challenging, with aggressive flight paths and rapidly moving cloud cover, leading to increased false positives. Dataset 5: Provided by \cite{vo2016spatiotemporal}, the ground truth accuracy is unknown.

\textbf{The Det-Fly dataset \cite{zheng2021air}} consists of 13,271 images of flying UAV targets (DJI Mavic) captured by the DJI M210 UAV . Compared to existing air-to-air datasets, this dataset is more systematic and comprehensive, covering a diverse range of real-world scenarios, including various backgrounds such as clear skies, mountains, fields, and urban areas, different viewpoints, relative distances (ranging from 10 meters to 100 meters), and flight altitudes (from 20 meters to 110 meters), as well as varying lighting conditions (captured across different times of the day). Additionally, the dataset includes challenging scenarios such as strong lighting, motion blur, and partial occlusion of targets.

\textbf{The MIDGARD dataset \cite{walter2020training}} is generated based on the DJI F550 frame, equipped with a Pixhawk flight controller and an Intel NUC computer, supporting both manual and autonomous control. The dataset covers a variety of indoor and outdoor environments, including forests, grasslands, fields, urban infrastructure, and complex indoor settings. The designed trajectories in the dataset include challenging scenarios such as target occlusion and line-of-sight loss.

\textbf{The Real World Dataset \cite{pawelczyk2020real}} is a new object detection dataset specifically designed for training computer vision-based object detection machine learning algorithms for binary object detection tasks. The dataset enables automated detection of multiple UAV objects using camera feeds from industrial cameras. It utilizes both publicly available UAV videos and videos recorded by the authors, extracting every 50 frames as data sources. The training dataset contains 51,446 images, and the test dataset contains 5,375 images. The dataset features diverse backgrounds, including urban, natural, and airport environments.

\textbf{The TIB-Net dataset \cite{sun2020tib}} was collected at Nanjing University of Aeronautics and Astronautics, containing 2,860 UAV images captured in real-world scenarios. It includes both multirotor and fixed-wing UAV with a resolution of 1920×1080 pixels. The images were taken with a fixed camera on the ground, approximately 500 meters away from the UAV, covering various lighting conditions from day to night. Each image is annotated with bounding boxes in the Pascal VOC format, and it includes challenging samples such as tiny, blurred UAV, and complex environments. 75\% of the data is used for the training set, and 25\% is used for the test set.

In conclusion, examples of the aforementioned datasets are shown in Figure \ref{newFig3}, detailed information is provided in Table \ref{table1}, and dataset access links can be found in Table \ref{table2}.

\section{ANTI-UAV Methods}

\begin{figure}[h]
\centering
\includegraphics[width=0.89\linewidth]{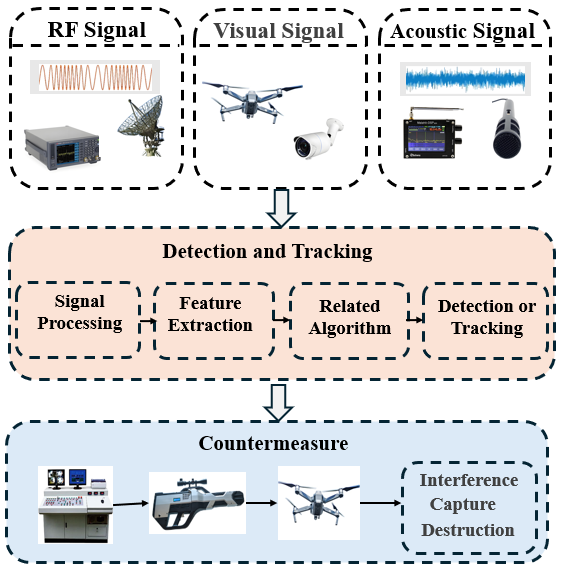}
\caption{A General Review of Anti-UAV Detection and Tracking Methods in Anti-UAV Systems.}
\label{newFig4}
\end{figure}

As shown in Figure \ref{newFig4}, Anti-UAV detection and tracking technologies have evolved through distinct stages. In the early stage, these systems primarily relied on single-sensor approaches, utilizing traditional radar \cite{ma2021method} and radio frequency \cite{huang2019unmanned}  jamming. However, these methods have a high false alarm rate due to problems such as weak reflection of small UAV, much environmental interference, overlapping signal bands, and reliance on active signals \cite{liu2020acoustic,aldowesh2019slow}.
In the mid-stage, with the proliferation of consumer UAV, countermeasure technologies gradually incorporated acoustic \cite{busset2015detection}, optical \cite{zhao2023anchor}, and infrared sensors \cite{xie2024rf}, attempting multi-sensor fusion \cite{nie2021uav}. However, the level of intelligence remained relatively low.
At the current stage, intelligent processing and multi-source fusion have been realized, leveraging  radar \cite{shao2022radar}, deep learning \cite{isaac2021unmanned}, and heterogeneous data fusion \cite{jiang2021anti} to enhance UAV detection accuracy and real-time response capabilities \cite{svanstrom2021real}. The mainstream Anti-UAV detection and tracking approaches are primarily vision-based methods or hybrid approaches that integrate visual data \cite{liu2021real,zhu2021tph,yan2023uav}.
This section categorizes vision-based methods and classifies the latest techniques into two categories: 
UAV-based Detection methods and UAV-based Tracking methods.

\subsection{UAV-based Detection}
To tackle the challenge of small UAV target detection in complex scenarios, researchers have built a lightweight and high-precision detection framework based on the YOLO series models through architectural innovation and multi-technique integration \cite{redmon2016you,wang2022fast,xiao2023fine,cheng2024anti,li2023global}.
At the methodological level, restructured network components significantly reduce model complexity while enhancing the ability to capture small target features \cite{gao2024lightweight,tian2024nfe,zhang2024ldhd,zhang2025bra}. The integration of attention mechanisms and dynamic gradient optimization strategies effectively improves resistance to interference in complex backgrounds \cite{gao2024novel,yang2025istd}. To address limitations in data availability and localization accuracy, GAN-based data augmentation \cite{zhou2025lightweight} is employed to improve the precision of small object localization.
Although these methods balance real-time performance and robustness, the detection of ultra-small targets in extreme environments remains challenging, necessitating further exploration of cross-modal fusion and fine-grained feature enhancement techniques.

To address the issue of insufficient anti-interference capability in UAV detection, Gao et al. \cite{gao2024novel} proposed an improved YOLOv9-based detection method, DotD-YOLOv9-C, and constructed a multi-category anti-interference dataset, Anti2, containing 5,062 images with 5,828 annotated bounding boxes. This dataset includes four target categories UAV, airplanes, helicopters, and birds and covers diverse backgrounds such as sky, ground, and clouds, along with small-object scenarios. The proposed method introduces Dot Distance as a positive-negative sample assignment metric, leveraging normalized Euclidean distance of center points to mitigate the sensitivity of traditional IoU-based localization for small objects. Furthermore, the method integrates YOLOv9-C’s GELAN network, which enhances gradient path diversity, and the PGI auxiliary monitoring framework, which provides reliable gradients through a reversible branch. Extensive experiments on the Anti2 dataset demonstrate that the method achieves 92.6\% mAP, showcasing its effectiveness in UAV detection under challenging conditions.

Awan et al. \cite{awan2025uett4k} proposed a UAV detection method based on the YOLOv6v3 model, aiming to address the challenge of detecting small UAVs in high-resolution  images. YOLOv6v3, introduced by Li et al. \cite{li2023yolov6}  , is a single-stage object detection framework selected for its balance between real-time performance and accuracy. The method enhances feature extraction through an improved BiC module and the SimCSPSPPF structure. It also employs Distributed Data Parallel  and TensorRT to accelerate training and inference. Experimental results show that the YOLOv6-L6 model achieved a mean Average Precision  of 92.7 percent on 4K images, while YOLOv6-S6 offered a better trade-off between speed and accuracy, making it more suitable for real-time applications.

Liu et al. \cite{liu2025design} proposed a UAV detection network based on deep feature fusion to improve small object detection performance in complex urban environments by optimizing the YOLOv8 model. The key contributions include the c-MPDIoU loss function for enhancing small object regression, the RFCAConv module for improving feature representation using receptive field and coordinate attention, and the MSI-FPN structure with EMA attention for better multi-scale feature fusion. Replacing the 20×20 detection head with a 160×160 head further improves small object perception. The model achieved an AP of 93.4 percent on the City-UAV dataset, outperforming YOLOv8 by 2.6 percent, and demonstrated strong robustness on both infrared and visible light UAV datasets.

Gao et al. \cite{gao2024lightweight} proposed a lightweight Anti-UAV detection method based on an improved YOLOv11, addressing the challenge of recognizing small UAV at long distances. The authors introduced HWD (Haar Wavelet Downsampling) as the downsampling module in the backbone network, reducing feature loss during extraction while significantly decreasing the number of parameters. To enhance small-object detection and improve scale adaptability, they replaced the original neck structure with a lighter CCFM (Cross-Channel Feature Modulation) architecture. Additionally, they removed the large-scale detection head and incorporated a new small-scale detection head, significantly improving the model’s ability to detect small UAV targets. Experimental validation on the DUT ANTI-UAV dataset demonstrated that, compared to the baseline YOLOv11 model, the improved version achieved 4\% higher precision, 4.5\% higher recall, 4.1\% improvement in mAP50, and 4.9\% improvement in mAP50-95, while reducing the number of parameters by 38.4\%. Although inference speed experienced a slight decline, the overall performance was significantly enhanced.

Tian et al. \cite{tian2024nfe} proposed a lightweight and efficient UAV detection network, named NFE-YOLO, designed to address the detection challenges of Low-altitude, Slow-speed, and Small-sized  UAV. Based on YOLOv8n, the network incorporates several key improvements: First, it introduces the EOrthoNet positive traffic attention mechanism, which enhances feature extraction through orthogonal compression filtering, thereby improving focus on small targets. Second, it replaces the original C2f structure with the C3Faster module, integrating partial convolution to reduce computational redundancy, bringing the model’s parameter size down to 3.72MB. Lastly, the neck network structure is optimized by adding a shallow small-object detection head and removing the large-object detection head, improving multi-scale feature fusion for better UAV detection performance.

Zhang et al. \cite{zhang2024ldhd} proposed a lightweight algorithm, LDHD-Net, designed to enhance the detection performance of small UAV in complex backgrounds. The authors introduced a novel Ghost-Shuffle feature extraction module, which removes redundant layers to reduce model complexity and incorporates a multi-scale feature mapping strategy to enhance target region representation. Additionally, a small-object detection structure was integrated into the shallow layers to extract richer UAV features, and a parallel dual-branch attention detection head was developed to address the differences between classification and regression tasks, thereby improving both detection accuracy and real-time capability. Experimental results demonstrate that LDHD-Net achieves superior detection performance and robustness across various challenging scenarios. However, the study also highlights certain challenges and limitations in real-world applications and suggests potential directions for future improvements.

Zhang et al. \cite{zhang2025bra} proposed an improved UAV small-object detection method, BRA-YOLOv10, based on YOLOv10, to address challenges in detecting UAV with low pixel ratios and high susceptibility to interference in complex backgrounds. The model introduces a Bidirectional Routing Attention  mechanism to enhance key feature focus while suppressing background noise. Additionally, a Small Target Detection Layer is designed to improve fine-detail extraction, and the SimCSPSPPF module, which integrates multi-scale pooling and cross-stage feature fusion, is employed to retain low-level feature information effectively. Experimental results demonstrate that BRA-YOLOv10 outperforms YOLOv10 across multiple key metrics, particularly in small-object detection rate, showing significant performance gains. Furthermore, its strong generalization ability across different datasets validates its potential for real-world UAV detection applications.

To address the limitations of existing deep learning methods in complex backgrounds, including target loss, high false detection rates, and insufficient real-time performance, Yuan et al. \cite{yang2025istd} proposed ISTD-DETR, an enhanced real-time detection Transformer model incorporating super-resolution preprocessing. This approach utilizes EDSR super-resolution to refine image details and employs data augmentation to improve generalization. Built upon the RT-DETR framework, the model integrates a state-space model and efficient multi-scale attention to optimize the backbone network for superior long-range dependency modeling. The feature fusion layer incorporates an SPD-EMA module, leveraging spatial-to-depth transformation and attention mechanisms to minimize critical information loss. Additionally, the introduction of S2 shallow features and a micro-target detection head enhances multi-scale feature fusion and localization accuracy. Experimental results demonstrate that ISTD-DETR achieves 96.2\% and 96.4\% mAP50 on the Anti-UAV and SIRST datasets, respectively, outperforming YOLO variants, DETR-based models, and segmentation methods while maintaining real-time processing at 133 FPS. This method presents a novel solution for high-precision, efficient infrared small-target detection in complex scenarios, offering both theoretical innovation and practical engineering value.

Zhou et al. \cite{zhou2025lightweight} proposed a lightweight UAV detection method, LAMS-YOLO, based on an improved YOLOv11, addressing challenges such as high model complexity and insufficient small-object feature extraction in complex environments. The method optimizes the backbone network by introducing depthwise separable convolutions and the H-Swish activation function, while integrating a linear attention mechanism with an LSTM-based gating system to enhance key feature focus. Additionally, an adaptive downsampling module, leveraging dynamic convolution and multi-scale fusion, is designed to achieve efficient feature compression. The model also incorporates an improved bounding box loss function to enhance localization accuracy. Experimental results demonstrate superior detection confidence and robustness against interference in complex backgrounds, small-object scenarios, and dense environments. However, challenges remain in extreme weather conditions and ultra-small UAV detection.

Phat et al. \cite{phat2025gan} proposed the GAN-UAV-YOLOv10s model to address the challenge of infrared small UAV detection in complex mountainous backgrounds. The main contribution of this work lies in the structural improvements to the YOLOv10s network, specifically enhancing the Neck and Head components to improve small-object detection capability. Moreover, it is the first study to integrate Generative Adversarial Networks for data augmentation in small UAV target detection, significantly boosting model accuracy. Experimental results demonstrate that this model outperforms multiple YOLO versions. Overall, by combining network optimization and GAN-based data augmentation, this study effectively enhances small UAV detection performance in mountainous environments.

Zhao et al. \cite{zhao2024lightweight} proposed a lightweight and efficient UAV small-object detection model, YOLOv8-E, which improves upon YOLOv8n in three key aspects: (1) The backbone network integrates the C2f-ESCFFM module, combining a SobelConv edge extraction branch with a standard convolution branch to enhance spatial information perception; (2) The neck network incorporates the CAHS-FPN structure, introducing a Context Anchor Attention mechanism to strengthen deep and shallow feature fusion, improving multi-scale detection performance; (3) The detection head adopts the LSCOD lightweight module, integrating Detail Enhancement Convolution and a shared convolution strategy, reducing the parameter count by over 50\%. Visualization heatmaps demonstrate that the improved model precisely focuses on the UAV’s overall structure while suppressing background interference, significantly reducing false detections in complex environments.

\subsection{UAV-based Tracking}
Anti-UAV tracking technologies, driven by the deep integration of multimodal approaches and deep learning, address challenges such as small target ambiguity, dynamic backgrounds, occlusion and multi-sensor coordination \cite{xie2024fora,xing2023uav,elleuch2024leveraging,zhang2023modality}. Three core technical pathways have emerged: Siamese-based, Self-attention-based and Vision fusion-based.
\subsubsection{\textbf{Siamese-based}}
The Siamese network framework achieves UAV tracking through deep feature matching \cite{li2019siamrpn++,hu2023siammask,xu2020siamfc++}. The initial frame extracts template features from the UAV region, while subsequent frames encode multi-scale search regions around predicted positions. Cross-correlation between template and search features generates a similarity map, with peak responses locating targets and a regression network refining bounding boxes. A dynamic template update strategy \cite{fang2021real,voigtlaender2020siam}  adapts to appearance changes. Motion prediction narrows search areas, and local re-detection handles occlusions, ensuring robustness in complex scenarios. Researchers \cite{huang2023anti,qian2025siamuf,wang2024contrastive} have enhanced feature extraction and localization accuracy by optimizing tracking network architectures, introducing innovative modules, and adopting dynamic template update strategies. They improve multi-scale feature correlation through global instance search, refine fine-grained feature representation by integrating hybrid attention mechanisms, and boost target localization robustness with background interference suppression components.
This Siamese-based method enables real-time UAV localization and trajectory analysis for single-target tracking.

Cui et al. \cite{cui2025strongsiamtracker} proposed a joint tracking method named StrongSiamTracker to enhance the robustness of anti-UAV tracking in infrared scenarios. This method dynamically integrates a global detection branch and a local Siamese tracking branch. It optimizes the YOLOv8s detector for small target detection by incorporating a dynamic detection head  and a loss function based on normalized Wasserstein distance(NWD). In the tracking branch, a Spatial Constraint Proposal Suppression module and an Adaptive Kalman Filter are introduced, along with a dynamic switching mechanism to address background interference and temporary target loss. Experiments show that the method achieved fourth and third place in Track 1 and Track 2 of the Fourth Anti-UAV Challenge \cite{4thAntiUAV}, respectively. On the Anti-UAV410 \cite{huang2023anti}  dataset, it reached a tracking accuracy of 0.6673, representing a 3.92 percent improvement over the baseline and ranking as the best-performing method among those that do not require an initial bounding box. Ablation studies confirm that the global detection branch contributes the most, while the dynamic detection head and NWD loss significantly improve the detector’s recall and mean average precision.

Zhang et al. \cite{zhang2025dlst} proposed DLST, a dual-branch learnable Siamese tracking framework designed for robust long-term tracking of small UAV targets in complex environments. It introduces a dual-template co-evolution mechanism that combines static and dynamic templates to handle deformation and occlusion. The framework includes DT-RPN for adaptive candidate generation, DT-RCNN for refined localization, and DT-BS for background suppression. With multi-stage training to reduce prediction errors, DLST outperforms previous methods on the Anti-UAV410 \cite{huang2023anti} dataset, achieving a success average of 68.81 percent. Ablation studies highlight the significant contributions of the dual-template design and DT-BS module.

Qin et al. \cite{qin2025pptracker} proposed PPTracker, a UAV swarm tracking framework that addresses challenges like high target similarity, dense occlusion, and scale variation by integrating historical tracking information through a prior-guided prompting mechanism. It uses a YOLOv11-based detection head with attention-guided feature enhancement and a Bot-SORT tracking head that combines Kalman filtering, camera motion compensation, and a hybrid appearance-spatial metric. On the MOT dataset of the Fourth Anti-UAV Challenge \cite{4thAntiUAV}, PPTracker achieved a MOTA of 67.9 percent, outperforming the baseline by 7.8 percent and effectively reducing identity switches caused by occlusion and motion.

To address the challenges of tracking small UAV targets in infrared scenarios and performing micro-object tracking in dynamic and blurred backgrounds, Huang et al. \cite{huang2023anti} drew inspiration from the successful implementations of Siamese trackers such as SiamATL \cite{huang2021siamatl}, SiamFC \cite{bertinetto2016fully}, SiamRPN \cite{li2018high}, DaSiamRPN \cite{zhu2018distractor},         SiamBAN \cite{chen2020siamese}, and SiamCAR \cite{guo2020siamcar}. Building upon these foundations, 
Author proposed a Anti-UAV tracking algorithm named SiamDT. The SiamDT method enhances the tracker's discrimination capability in dynamic background clutter through a dual-semantic feature extraction mechanism. It consists of three key components: Dual-Semantic Region Proposal Network , Versatile Region Convolutional Neural Network , Background Interference Suppression.

Qian et al. \cite{qian2025siamuf} proposed SiamUF, a Siamese tracker based on a dense U-shaped network, as an improvement over SiamCAR. The proposed SiamUF incorporates the following key enhancements:
Dense Nested U-shaped Feature Network with Hybrid Attention Mechanism, which enhances the extraction of both fine-grained details and semantic features for infrared small target tracking.
Normalized Wasserstein Distance Regression Loss, which mitigates the severe IoU fluctuations of small target bounding boxes.
Side Window Filter, which suppresses background edge noise while preserving high-frequency information of the target.
Dynamic Template Update Mechanism, which adaptively integrates historical templates with current detection results to handle target deformation and occlusion.
Experimental results demonstrate that SiamUF achieves 83.1\% tracking accuracy (CLE $\leq$ 10 pixels) and real-time performance at 30 FPS on publicly available infrared datasets.

CAMTracker proposed by Wang et al. \cite{wang2024contrastive} is used to solve the problems faced by anti-UAV tracking in thermal infrared videos, such as sparse small target features, large background interference, and long-term tracking template feature degradation.
IG-RPN network solves the problem of small target positioning difficulties at low resolution through global instance search and template-search area feature correlation;
CAM module optimizes the loss function through contrastive learning to solve the problem of low discrimination between positive and negative samples caused by background noise interference;
ADM module adaptively updates the template through a dynamic memory mechanism to solve the feature drift caused by static templates in long-term tracking.
The combination of the three forms a collaborative optimization framework, which realizes efficient tracking of low-resolution thermal infrared videos on the Anti-UAV410 dataset and significantly improves the robustness to target scale changes and complex backgrounds.This method promotes the development of anti-UAV tracking in thermal infrared video and provides new ideas for anti-UAV tracking in complex situations in the future.

\subsubsection{\textbf{Self-attention-based}}

Self-attention-based Anti-UAV systems achieve precise target perception through global modeling and end-to-end optimization \cite{mayer2022transforming}. The architecture encodes image patches into sequential features, where the Transformer's self-attention mechanism captures global dependencies, effectively addressing small-object ambiguity and background interference \cite{gao2022aiatrack,lin2022swintrack}. The decoder employs learnable queries for direct bounding box prediction, eliminating anchor designs. For tracking, spatiotemporal attention maintains identity consistency across frames, while trajectory prediction handles motion patterns and occlusions \cite{tong2024st,qian2025transist}. By optimizing detection network architectures, introducing innovative components, and adopting dynamic learning strategies, researchers enhance global semantic perception through multi-scale feature modeling \cite{yang2025istd,yu2023unified}. They refine fine-grained feature representation by integrating spatial-to-depth transformation techniques while improving computational efficiency through depthwise separable convolution structures and lightweight designs.

Qian et al. \cite{qian2025transist} proposed a Self-attention-based infrared small target tracking algorithm called TransIST, which addresses the challenges of weak target features, strong edge interference, and complex background noise in UAV scenarios with complex sky backgrounds. By employing multi-scale feature modeling and a dynamic learning strategy, the tracking performance is enhanced. The multi-scale dilated attention (MSDA) is introduced into the Transformer backbone network to improve the perception of local details and global semantics of small targets. Additionally, a side window filter (SWF) is utilized to suppress the interference of edge noise on target localization, and an exponential moving average (EMA) learning strategy is adopted to dynamically smooth the network parameter update process, optimizing training stability and generalization performance. TransIST significantly outperforms traditional correlation filtering and existing deep learning methods in tracking accuracy and success rate on publicly available near-infrared datasets, achieving real-time performance at 30 FPS on a GPU platform. This method deeply integrates the Transformer architecture with the infrared small target tracking task, collaboratively addressing the issues of insufficient target features and edge interference through an end-to-end multi-module approach, providing an efficient solution for UAV countermeasures.

Yu et al. \cite{yu2023unified} proposed a unified UAV tracking approach called UTTracker, based on the Transformer architecture. To address challenges in thermal infrared video such as severe target deformation, frequent occlusion, camera motion, and small object tracking, they designed a multi-module collaborative framework comprising a local tracking module, global detection module, background correction module, and a dynamic small object module. The method incorporates a dynamic switching mechanism between local tracking and global detection, a cross-frame background alignment strategy, and multi-scale feature collaboration. It achieved an AUC score of 77.9\% on the Anti-UAV Challenge dataset \cite{4thAntiUAV}, surpassing baseline models like SiamSTA \cite{huang2021siamsta} by 5.3\%, and won second place in the 3rd Anti-UAV Challenge. Experimental results demonstrate that each module contributes significantly to performance improvement, though there remains room for enhancement in tracking extremely small objects under dynamic backgrounds.

\subsubsection{\textbf{Vision fusion-based}}

Compared to single-sensor approaches, vision-based fusion methods demonstrate distinct advantages in addressing critical challenges in cluttered environments, such as small target miss detection, trajectory abrupt changes, and single-sensor failure risks, by integrating multi-modal data to enhance detection robustness and tracking continuity while maintaining computational efficiency.
Vision-based fusion methods employ heterogeneous visual sensors—such as multispectral cameras, LiDAR, and stereo vision systems—to synchronously capture multimodal target data \cite{svanstrom2021real,cheng2023slbaf,yang2019uav}. Through coordinated spatiotemporal registration and feature alignment, deep learning models perform feature-level or decision-level fusion on multisource imagery, augmenting the discernibility of UAV targets’ contours, thermal signatures, and motion patterns \cite{chen2022multi,liu2024asifusion}. This framework dynamically integrates multi-perspective observations for predicting flight trajectories via adaptive fusion mechanisms. By leveraging heterogeneous data complementarity, the approachs \cite{xie2024rf,vitiello2024radar,wang2024rtm} effectively mitigates critical challenges in complex environments, including small-target omission, abrupt trajectory deviations, and single-modality sensor failures, while ensuring robust cross-modal consistency in detection and tracking performance.


Xie et al. \cite{xie2024rf} proposed a radio-frequency-vision fusion framework to address the challenge of precise positioning of small UAV in complex environments. To solve the limitations of traditional single-modal methods (such as radar's susceptibility to small target interference, vision's potential confusion with birds, and radio frequency's vulnerability to electromagnetic noise), the authors achieve time-space alignment between radio frequency signals and visual images through joint calibration of the array antenna and camera. They also use the array antenna to extract spatial spectra, focusing on target areas within the image modality. The authors innovatively introduce a segmentation-based denoising method that effectively removes spectral noise and enhances the joint calibration accuracy. Furthermore, they developed an Anti-UAV positioning experimental platform and constructed a dataset containing synchronized visual images and radio frequency signals. Experiments demonstrate that the framework, by leveraging the complementary nature of heterogeneous modalities, outperforms traditional methods in positioning accuracy and interference resistance, providing a high-reliability real-time positioning solution for Anti-UAV systems.

\begin{table*}
\centering
\renewcommand{\arraystretch}{1.5}
\caption{Detailed description of the datasets used, experimental results, and experimental environments for different methods.}
\label{table3} 
\resizebox{\textwidth}{!}{%
\begin{tabular}{cccll} 
\hline
\textbf{Category} & \textbf{Method} & \textbf{Dataset} & \multicolumn{1}{c}{\textbf{Experimental Result}} & \multicolumn{1}{c}{\textbf{Experimental Environment}} \\ 
\hline
\multirow{22}{*}{UAV-based Detection} 
& DotD-YOLOv9-C~\cite{gao2024novel} & \begin{tabular}[c]{@{}c@{}}Self-built dataset\\Anti2\end{tabular} & mAP=92.60\% & \begin{tabular}[c]{@{}l@{}}NVIDA GeForce RTX 4060, Ubuntu 18.04,\\PyTorch 2.3.0, Python 3.8.19\end{tabular} \\ 
\cline{2-5}
& YOLOv6v3-Based~\cite{awan2025uett4k} & \begin{tabular}[c]{@{}c@{}}Self-built dataset\\UETT4K Anti-UAV\end{tabular} & \begin{tabular}[c]{@{}l@{}}Precision=96.2\%, Recall=90.0\%, F1=93.5\%\\mAP50=92.7\%, mAP50-95=70.3\%, FPS=36.4\end{tabular} & \begin{tabular}[c]{@{}l@{}}sys-220GP-TRN server with dual A30 GPUs (24GB+26GB),\\256GB memory, Intel Xeon Gold 6354, PyTorch\end{tabular} \\
\cline{2-5}
& c-MPDIoU~\cite{liu2025design} & \begin{tabular}[c]{@{}c@{}}City-UAV, Anti-UAV~\cite{jiang2021anti},\\DUT ANTI-UAV~\cite{zhao2022vision}\end{tabular} & \begin{tabular}[c]{@{}l@{}}City-UAV AP=93.4\%\\Anti-UAV AP=88.1\%\\DUT ANTI-UAV AP=96.7\%\end{tabular} & \begin{tabular}[c]{@{}l@{}}RTX 3090, Intel Xeon Silver 4210R,\\PyTorch, CUDA 12.1, Ubuntu 20.04\end{tabular} \\
\cline{2-5}
& YOLOv11-Based~\cite{gao2024lightweight} & DUT ANTI-UAV~\cite{zhao2022vision} & \begin{tabular}[c]{@{}l@{}}Precision=96.9\%, Recall=89.8\%,\\mAP50=95.2\%, mAP50-95=65.7\%, FPS=91\end{tabular} & \begin{tabular}[c]{@{}l@{}}Ubuntu 22.04, RTX 3090, PyTorch\end{tabular} \\
\cline{2-5}
& NFE-YOLO~\cite{tian2024nfe} & UESTC Anti-UAV & \begin{tabular}[c]{@{}l@{}}Precision=98.3\%, Recall=96.1\%,\\mAP50=98.7\%, mAP75=90.8\%\end{tabular} & \begin{tabular}[c]{@{}l@{}}Ryzen 5 3400G, 16GB RAM, Tesla T4 (×2),\\PyTorch1.11.0, Torchvision0.11.0\end{tabular} \\
\cline{2-5}
& LDHD-Net~\cite{zhang2024ldhd} & Anti-UAV410, SIDD~\cite{yuan2023irsdd} & AP=95.6\% & \begin{tabular}[c]{@{}l@{}}Windows 11, PyTorch 1.13.0,\\Torchvision 0.14.0, Python 3.8.15, CUDA 11.6.1\end{tabular} \\
\cline{2-5}
& BRA-YOLOv10~\cite{zhang2025bra} & SIDD~\cite{yuan2023irsdd} & \begin{tabular}[c]{@{}l@{}}Precision=98.9\%, Recall=92.3\%,\\mAP50=96.5\%, mAP50-95=67.6\%\end{tabular} & \begin{tabular}[c]{@{}l@{}}RTX 4090, PyTorch 2.0.1, Python 3.9.19,\\Ubuntu 18.04, CUDA 11.7\end{tabular} \\
\cline{2-5}
& ISTD-DETR~\cite{yang2025istd} & Anti-UAV410, SIRST & \begin{tabular}[c]{@{}l@{}}SIRST: Precision=94.8\%, Recall=94\%, mAP50=96.4\%\\Anti-UAV410: mAP50=96.2\%, mAP50-95=55.3\%, FPS=133.1\end{tabular} & \begin{tabular}[c]{@{}l@{}}Ubuntu 22.04.1, RTX A5000\end{tabular} \\
\cline{2-5}
& LAMS-YOLO~\cite{zhou2025lightweight} & Self-built dataset & \begin{tabular}[c]{@{}l@{}}F1=92.58\%, Precision=96.10\%,\\Recall=86.30\%, mAP30=93.4\%, mAP95=74.7\%\end{tabular} & \begin{tabular}[c]{@{}l@{}}Ubuntu 20.04, Tensor Core A100,\\Python 3.9.11, PyTorch 1.10.0\end{tabular} \\
\cline{2-5}
& GAN-UAV-YOLOv10s~\cite{phat2025gan} & SIDD & Accuracy=96\%, Recall=84\%, mAP50=92\% & \begin{tabular}[c]{@{}l@{}}RTX3090, Intel E5-2667v, PyTorch 1.13.1,\\Python 3.10.13\end{tabular} \\
\cline{2-5}
& YOLOv8-E~\cite{zhao2024lightweight} & Real ward~\cite{pawelczyk2020real} & \begin{tabular}[c]{@{}l@{}}Precision=98.5\%, Recall=95.8\%,\\mAP=98.4\%, mAP50-95=72.1\%, FPS=57.4\end{tabular} & \begin{tabular}[c]{@{}l@{}}Windows 10, GTX 1050 Ti, CUDA 11.8,\\Python 3.9.19, PyTorch 2.0.0\end{tabular} \\
\hline
\multirow{17}{*}{UAV-based Tracking} 
& StrongSiamTracker~\cite{cui2025strongsiamtracker} & 4th-Anti-UAV, Anti-UAV410 & Anti-UAV410 acc=66.73\% & \begin{tabular}[c]{@{}l@{}}2× RTX 4090, Python 3.7, PyTorch 1.8.1\end{tabular} \\
\cline{2-5}
& DLST~\cite{zhang2025dlst} & Anti-UAV410 & SA=68.81\% & RTX 4090, PyTorch \\
\cline{2-5}
& PPTracker~\cite{qin2025pptracker} & 4th-Anti-UAV & MOTA=67.9\% & 2× RTX 3090, PyTorch \\
\cline{2-5}
& SiamDT~\cite{huang2023anti} & \begin{tabular}[c]{@{}c@{}}Self-built\\Anti-UAV410\end{tabular} & SA=68.19\% & Not provided \\
\cline{2-5}
& SiamUF~\cite{qian2025siamuf} & \begin{tabular}[c]{@{}c@{}}Public datasets\\\cite{sun2021dataset}, \cite{hui2020dataset}\end{tabular} & \begin{tabular}[c]{@{}l@{}}Precision=83.07\%, AUC=65.66\%, FPS=30\end{tabular} & RTX3090, PyTorch, Python \\
\cline{2-5}
& CAMTracker~\cite{wang2024contrastive} & Anti-UAV410 & \begin{tabular}[c]{@{}l@{}}Precision=88.56\%, AUC=66.68\%, SA=67.10\%\end{tabular} & \begin{tabular}[c]{@{}l@{}}i5-13600KF, RTX4090, PyTorch 1.13,\\CUDA 11.7\end{tabular} \\
\cline{2-5}
& TransIST~\cite{qian2025transist} & \begin{tabular}[c]{@{}c@{}}Public: \cite{sun2021dataset}, \cite{hui2020dataset}, \cite{xiong2020material}\end{tabular} & \begin{tabular}[c]{@{}l@{}}AUC=66.0\%, Accuracy=88.1\%, FPS=30\end{tabular} & Not provided \\
\cline{2-5}
& UTTracker~\cite{yu2023unified} & \begin{tabular}[c]{@{}c@{}}1st \& 2nd Anti-UAV~\cite{4thAntiUAV}\end{tabular} & \begin{tabular}[c]{@{}l@{}}1st: AUC=77.9\%, Precision=98.0\%\\2nd: AUC=72.4\%, Precision=93.4\%\end{tabular} & \begin{tabular}[c]{@{}l@{}}4× RTX 3090, Python 3.6, PyTorch 1.7.1\end{tabular} \\
\cline{2-5}
& ISD-UNet~\cite{xie2024rf} & Self-built dataset & TCR=96.4\%, AP=44.7\% & \begin{tabular}[c]{@{}l@{}}GTX (24GB), AMD EPYC 7302@3.0GHz,\\PyTorch1.12.1, Python 3.9, Ubuntu 18.04.6\end{tabular} \\
\cline{2-5}
& Fuse-Before-Track~\cite{vitiello2024radar} & Self-built dataset & \begin{tabular}[c]{@{}l@{}}RMS error: within meters (distance),\\below 1 m/s (speed)\end{tabular} & \begin{tabular}[c]{@{}l@{}}Echoflight MESA, FLIR Blackfly,\\Ublox LEA-6T, M8T, DJI M100\end{tabular} \\
\cline{2-5}
& RTM-UAVDet~\cite{wang2024rtm} & Anti-UAV~\cite{jiang2021anti} & AP=66.1\%, FPS=72.4 & \begin{tabular}[c]{@{}l@{}}RTX 3090, Python 3.8, Torch 1.7.1\end{tabular} \\
\hline
\end{tabular}
}
\end{table*}

Vitiello et al. \cite{vitiello2024radar} proposed a radar-vision fusion framework using a "Fuse-Before-Track" strategy to solve single-sensor limitations in non-cooperative UAV sensing. Their two-stage approach integrates CNN-based visual detection to filter radar echoes at the detection level, suppressing clutter, and combines radar ranging with vision-derived angles at the tracking level via a nonlinear constant velocity model and Kalman filtering, enhancing azimuth estimation. Validated by centimeter-level GNSS benchmarks, the method achieves meter-level radar ranging accuracy post-detection fusion and reduces horizontal distance errors by over 50\% through threat assessment in tracking fusion, offering an all-weather solution for low-altitude autonomous obstacle avoidance.

Wang et al. \cite{wang2024rtm} proposed a real-time multi-modal UAV detection framework, RTM-UAVDet, to address the complementary detection needs of visible light and thermal infrared images under extreme conditions . The method overcomes the limitations of traditional multi-modal fusion approaches that rely on image alignment and are computationally complex. By designing a multi-modal dynamic convolution model based on Convolutional Neural Networks, the framework is capable of parallelly extracting features from both RGB and TIR modalities and enabling cross-modal interaction without the need for pre-alignment, allowing it to process static multi-modal images. Experimental results show that the framework achieves an average precision of 66.1\% in validation tasks, with a real-time performance of 72.4 frames per second, effectively detecting UAV targets with a size of over 48 pixels. This approach significantly enhances detection robustness in complex environments by combining the high-resolution details of visible light with the all-weather perception capabilities of thermal infrared, providing an efficient real-time multi-modal solution for scenarios such as public safety and airspace monitoring.

\subsection{Evaluation Metrics}
According to the methods discussed in this paper, different approaches employ distinct evaluation metrics.  True Positive (TP), which represents the number of correctly detected targets, False Positive (FP), which refers to the number of background regions mistakenly classified as targets, True Negative (TN), which denotes the number of correctly predicted negative samples, and False Negative (FN), which indicates the number of targets misclassified as background. In addition, a comprehensive evaluation index State Accuracy (SA) is proposed in this paper. The evaluation indicators in Table \ref{table3} are as follows.

The State Accuracy (SA) \cite{huang2023anti}  metric evaluates tracking performance by combining target localization accuracy and presence detection. The formula is defined as:
\small
\begin{equation}
SA = \frac{1}{N} \sum_{t=1}^{N} \left( IoU_t \times \delta(v_t > 0) + p_t \times (1 - \delta(v_t > 0)) \right),
\label{eq:state_accuracy}
\end{equation}
where State Accuracy metric incorporates both tracking precision and presence detection through its components, $N$ represents the total number of samples, while $t$ indexes individual samples. The term $IoU_t$ denotes the Intersection-over-Union score measuring localization accuracy for the $t$-th sample. The indicator function $\delta(v_t > 0)$ evaluates to 1 when the ground-truth visibility flag $v_t$ indicates a visible target (i.e., $v_t > 0$), and 0 otherwise. The prediction term $p_t$ reflects the system's presence detection, taking value 1 when the target is predicted as absent (empty prediction) and 0 when present. This formulation effectively balances the evaluation between accurately tracking visible targets (first term) and correctly identifying absent targets (second term).


The tracking accuracy ($\text{acc}$) \cite{4thAntiUAV} consists of two components. The first component is the average score over all frames. When the target is visible ($v_i > 0$, i.e., $\delta(v_i > 0) = 1$), the score is computed as the Intersection over Union (IoU) between the predicted bounding box and the ground-truth box. When the target is not visible ($\delta(v_i > 0) = 0$), the score is determined by the predicted visibility flag $p_i$, where $p_i = 1$ indicates that the predicted box is empty. In this case, a correct invisibility prediction receives a score of 1, while an incorrect prediction receives a score of 0. The second component is a penalty term designed to penalize false negatives, calculated as the 0.3 power of the average proportion of frames in which the target is visible but the prediction is empty (i.e., $p_i \cdot \delta(v_i > 0)$), and multiplied by 0.2. The final accuracy is obtained by subtracting the penalty term from the first component. Here, $T$ represents the total number of frames, and $T^*$ denotes the number of frames where the target is present in the ground truth, defined as $T^* = \sum_{i=1}^{T} \delta(v_i > 0)$.

\begin{align}
\text{acc} =\ & \frac{1}{T} \sum_{t=1}^{T} \left( \text{IoU}_t \cdot \delta(v_t > 0) 
+ p_t \cdot (1 - \delta(v_t > 0)) \right) \notag \\
& - 0.2 \cdot \left( \frac{1}{T^*} \sum_{t=1}^{T^*} p_t \cdot \delta(v_t > 0) \right)^{0.3}.
\end{align}

MOTA (Multi-Object Tracking Accuracy) \cite{4thAntiUAV} is a metric used to evaluate the overall performance of multi-object tracking algorithms. 
It combines the effects of false positives , false negatives , and ID switches (IDS). 
$\text{GT}$ represents the total number of ground-truth objects.

MOTA ranges from 0 to 1, with higher values indicating better tracking accuracy. 
It is a comprehensive metric that reflects tracking performance by considering these types of errors.
\begin{equation}
\text{MOTA} = 1 - \frac{\text{FP} + \text{FN} + \text{IDS}}{\text{GT}}.
\end{equation}

Accuracy \cite{phat2025gan} measures the proportion of correct predictions, including both positive and negative classes, among all predictions made by the model.

\begin{equation}
    \text{Accuracy} = \frac{TP + TN}{TP + FN + FP + TN}.
    \label{eq:accuracy}
\end{equation}

The accuracy measure \cite{yu2023unified} of the proportion of samples predicted as positive that are actually true positive can be defined as follows:
\begin{equation}
    \text{Precision} = \frac{TP}{TP + FP}.
    \label{eq:precision}
\end{equation}

Recall \cite{zhao2024lightweight} measures the proportion of actual positive samples that are correctly predicted as positive by the model, and its definition is as follows:
\begin{equation}
    \text{Recall} = \frac{TP}{TP + FN}.
    \label{eq:recall}
\end{equation}


The F1 score \cite{zhou2025lightweight} is the harmonic mean of precision and recall, providing a balanced measure that considers both false positives and false negatives. It is defined as follows:
\begin{equation}
    F1 = \frac{2 \cdot Precision \cdot Recall}{Precision + Recall}.
    \label{eq:f1}
\end{equation}

FPS (Frames Per Second) \cite{gao2024lightweight} represents the number of frames processed per second, where t is the processing time per frame:
\begin{equation}
    FPS = \frac{1}{t}.
    \label{eq:fps}
\end{equation}

Average Precision(AP) \cite{zhang2024ldhd} is the area under the Precision-Recall curve (AUC), used to evaluate the overall performance of the model for a single class.
\begin{equation}
    AP = \int_{0}^{1} \text{Precision}(Recall) \, d(Recall),
    \label{eq:ap}
\end{equation}
where the integration range spans from a recall value of 0 to 1.

The TCR (Target Coverage Rate) \cite{xie2024rf} formula is used to evaluate the effectiveness of cross-modal fusion methods in small target localization.
\begin{equation}
    TCR = \frac{N_t}{N_s} \times 100\%,
    \label{eq:tcr}
\end{equation}
where \( N_t \) represents number of slices containing the target, \( N_s \) denotes total number of slices. Slices refer to localized regions or subunits divided from input data to facilitate block-by-block analysis.

Mean Average Precision (mAP) \cite{gao2024novel} is the average of the Average Precision (AP) values across all classes, defined as:
\begin{equation}
    \text{mAP} = \frac{1}{N} \sum_{i=1}^{N} \text{AP}_i,
    \label{eq:map}
\end{equation}
where
\( N \) is the total number of classes,
\( \text{AP}_i \) denotes the Average Precision for the \( i \)-th class.
As a core metric in object detection tasks, mAP evaluates the model's overall performance across multiple object categories.

Through a systematic analysis of evaluation metrics including SA, Accuracy, Precision, Recall, F1, FPS, AP, TCR, and mAP, we observe that mAP has become the universal benchmark due to its comprehensive assessment of multi-class detection accuracy \cite{gao2024lightweight,tian2024nfe}. Precision and Recall remain indispensable for binary classification tasks requiring high sensitivity \cite{zhang2025bra,zhou2025lightweight}, while FPS gains significant attention in engineering deployments for its direct reflection of real-time performance \cite{yang2025istd,zhao2024lightweight,qian2025transist}. The emerging SA metric demonstrates unique value in dynamic occlusion scenarios by jointly evaluating localization accuracy and presence detection \cite{huang2023anti}, whereas TCR specializes in cross-modal small target coverage analysis \cite{xie2024rf}. The selection of these metrics is fundamentally driven by task-specific requirements, while being jointly influenced by domain consensus and technological trends.

\subsection{Discussion of Method Results}

From Table \ref{table3}, we can see UAV-based Detection methods \cite{gao2024novel,phat2025gan,zhang2024ldhd,zhang2025bra,zhou2025lightweight} adopt a one-stage detection network to achieve end-to-end object detection, striking a balance between speed and accuracy that makes them suitable for real-time monitoring scenarios, although their detection precision may suffer when dealing with small targets or complex backgrounds. Siamese-based methods \cite{huang2023anti,qian2025siamuf,wang2024contrastive} primarily construct a similarity metric between the target template and the search region, enabling rapid matching and tracking of the UAV's appearance features, and offering good real-time performance and robustness; however, they may encounter limitations when there are significant changes in the target's scale or pose. Self-attention-based approaches \cite{qian2025transist,yu2023unified} utilize self-attention mechanisms to capture global information and long-range dependencies, effectively addressing issues such as target occlusion and background interference to enhance overall detection and tracking performance; however, they often entail higher computational complexity, which may 
require further optimization to meet real-time requirements. Visual fusion-based methods \cite{xie2024rf,vitiello2024radar,wang2024rtm}, on the other hand, integrate multi-scale or multi-modal visual features to fully exploit the complementary information available from different sensors, thereby significantly improving target detection and tracking accuracy in challenging environments such as low light or harsh weather conditions, and providing robust support for precise UAV localization and stable tracking.


\section{Limitations and Future Discussion}
Through the investigation of datasets and methods, this section describes the existing limitations of current datasets and state-of-the-art methods. Additionally, seven potential future research directions are proposed.

\subsection{About Limitations of datasets}
Existing Anti-UAV detection and tracking datasets have played a crucial role in advancing Anti-UAV technology. However, they still have several limitations, which are mainly reflected in the following aspects:
\begin{itemize}  
 \item \textbf{Lack of Diversity. }Many datasets have limitations in terms of variations in environmental conditions, such as different lighting, weather, and background complexity, which directly affect the model's generalization ability \cite{gao2024novel,dai2023one,li2020reconstruction}. Specifically, when the training dataset lacks diverse environmental conditions, the model may overfit to specific scenes or conditions, resulting in poor performance in real-world applications \cite{svanstrom2021dataset}. For example, under varying lighting conditions, the model may fail to effectively recognize targets because it has not encountered similar lighting variations during training. Additionally, extreme weather conditions that affect visual sensors can also lead to a decline in the model's performance.
 
  \item \textbf{Lack of Multi-UAV Datasets. }Existing Anti-UAV detection and tracking datasets primarily focus on single-target scenarios, with limited datasets available for multiple UAV \cite{huang2023anti,4thAntiUAV}. As the prevalence of multi-UAV cooperative operations increases in the future, current single-target datasets are insufficient to support advanced research in this area.
  
  \item \textbf{Limited Multimodal Data. }Most datasets primarily focus on RGB images or thermal imaging images, with few providing synchronized multimodal data, which is crucial for achieving robust object detection in complex environmental conditions \cite{jiang2021anti,yuan2024mmaud,svanstrom2021dataset}. Specifically, single-modal data often fails to capture the full range of information about a target. For example, RGB images perform well in well-lit conditions, but their recognition capability can significantly degrade in low-light or high-glare environments \cite{zhao2022vision}. While thermal imaging has an advantage in low-light conditions, it can also face difficulties in recognition when the background temperature is high or the object surface temperature is close to the ambient temperature \cite{yuan2023irsdd}.
  
  \item \textbf{Inconsistent Labeling Standards. }Different datasets follow different annotation standards \cite{yuan2024mmaud,jiang2021anti}, which can create difficulties during model training and evaluation due to inconsistent definitions of ground truth values. Specifically, differences in annotation standards may be reflected in aspects such as target category classification, bounding box annotation methods, and the level of detail in labels \cite{zhao2022vision,hui2019dataset}. For example,  a target may be labeled as a broad category, while in others, it may be subdivided into multiple subclasses. Such discrepancies make it challenging for the model to learn consistent features during training, ultimately affecting its generalization ability.
  
  \item\textbf{Lack of Real-Time Benchmarking. }Many datasets do not include real-time constraints or latency benchmarks \cite{guo2024global,walter2020training} , making it difficult to evaluate the feasibility of detection and tracking algorithms in real-world deployment. Specifically, real-time constraints refer to the requirement that an algorithm must complete its tasks within a specific time frame to meet the demands of real-time applications. Without clear real-time performance metrics in the dataset, researchers cannot accurately assess the algorithm's response time and processing capability in practical applications \cite{pawelczyk2020real,sun2020tib}.
  Furthermore, latency benchmarks refer to the time an algorithm takes to process input data, including the total time for data acquisition, processing, and output. Without these benchmarks, developers may overestimate the performance of the algorithm, leading to excessive delays during actual deployment, which can affect the overall efficiency of the system.
\end{itemize}

\subsection{About Limitations of Methods}
Anti-UAV technology has been developed to detect, interfere with, or destroy potential UAV threats. However, existing Anti-UAV methods still have some limitations, mainly in the following aspects:
\begin{itemize}  
  \item \textbf{Balance Between Real-Time Performance and Accuracy.} 
  High-precision detection algorithms often struggle to meet real-time requirements due to their high computational complexity \cite{gao2024lightweight,yang2025istd,qian2025transist}. 
  For example, deep learning-based detection algorithms achieve excellent accuracy, but their complex model structures result in slow inference speeds, making real-time detection difficult. 
  Conversely, lightweight models can improve detection speed but often at the cost of reduced accuracy \cite{qian2025siamuf}. 
  This trade-off between accuracy and real-time performance makes it challenging to find an ideal balance in practical applications. 
  Although existing algorithms have made progress in handling large-scale datasets, they still face significant challenges in balancing real-time performance and accuracy.
  
  \item \textbf{Multi-target tracking.} 
  In multi\textendash UAV collaborative flight scenarios, tracking algorithms often face issues of target confusion or loss \cite{huang2023anti,4thAntiUAV,wang2024rtm}. 
  When multiple \texttt{UAV} appear in the image simultaneously, their appearance features are similar, and during motion, they are prone to mutual occlusion or close proximity, making it difficult for the algorithm to accurately distinguish and continuously track each target. 
  In addition, existing tracking algorithms lack robustness against target scale variation, rapid movement, and background interference, further increasing the difficulty of multi-target tracking \cite{xie2024rf,vitiello2024radar}. 
  Although the introduction of deep learning and multi-target tracking technology has improved algorithm performance, target loss or mis-tracking issues still persist in complex scenarios.
  
  \item \textbf{Sensitivity to target appearance changes.} 
  In Siamese-based methods, the tracking relies on similarity matching between the target template and the search region \cite{wang2024contrastive,yu2023unified}. 
  Therefore, when there are significant changes in the target's scale, posture, or appearance, matching failure is likely to occur, resulting in instability in detection and tracking.
  
  \item \textbf{Complexity of multimodal data fusion.} 
  In multimodal data fusion, data acquired by different sensors may differ in terms of time, space, and scale \cite{vitiello2024radar}. 
  To achieve effective fusion, these data must be accurately aligned, which not only requires complex algorithmic support but may also involve real-time processing demands, thereby increasing the computational burden on the system \cite{huang2023anti,xie2024rf}. 
  Additionally, data from different modalities have distinct feature representations and information content. 
  Effectively extracting and fusing these features to enhance detection and tracking accuracy while maintaining information integrity is a challenging task \cite{yang2025istd,yu2023unified}. 
  Multimodal data fusion systems typically require the integration of multiple sensors and processing units, which not only increases hardware complexity but also places higher demands on the system's stability and reliability.
\end{itemize}

\subsection{About Prospects of Future} 
Anti-UAV detection and tracking technology will evolve in future research towards balancing real-time performance and accuracy, multimodal data fusion and heterogeneous sensor collaboration, insufficient generalization in diverse environments, target confusion and occlusion in multi-UAV scenarios,  model robustness and false detection suppression, modeling structural deformation and dynamic appearance
for UAV and legal, ethical, and privacy considerations. These advancements aim to better address the increasingly complex and diverse UAV threats. The following are potential directions for future research:

\begin{itemize}

    \item \textbf{Balancing real-time performance and accuracy.} To address the latency caused by the high computational complexity of precision algorithms, future work should focus on achieving low-latency, high-reliability detection and tracking \cite{wu2020real,fang2021real}. This can be accomplished by designing lightweight model architectures and introducing dynamic computation allocation strategies that adapt to scene complexity. Additionally, model compression techniques such as quantization, pruning, and knowledge distillation can effectively reduce computational demands while maintaining accuracy. These approaches are particularly valuable for real-time applications in border patrol and urban security.

    \item \textbf{Multimodal data fusion and heterogeneous sensor collaboration.} In the future, integrating multisource data such as infrared, visible light, radar, and acoustic signals\cite{kumawat2022diat,sharma2024advanced,dong2023drone,zhao2022vision} can greatly enhance target recognition, especially under adverse conditions. A cross-modal Transformer architecture may be employed to adaptively fuse heterogeneous features, though challenges remain in sensor synchronization and data alignment. Additionally, quantum sensing combined with quantum radar fusion\cite{torrome2024advances} offers ultra-sensitive detection capabilities by integrating technologies like atomic magnetometers and quantum illumination. This approach is particularly effective for tracking nano-scale UAVs in complex electromagnetic or extreme weather environments.
    
    \item \textbf{Insufficient generalization in diverse environments.} To enhance model adaptability in complex and variable scenarios, future research can focus on constructing virtual-real hybrid datasets that integrate challenging weather conditions and multi-UAV collaborative tasks, thereby enriching the diversity and representativeness of training data. However, achieving high-fidelity simulation and domain alignment between virtual and real data remains a technical challenge. Dynamic data augmentation strategies (e.g., \cite{kim2022exploring}, \cite{shen2022data}) can further enhance model robustness, yet may require adaptive mechanisms to avoid overfitting to synthetic variations. Moreover, integrating super-resolution techniques \cite{xu2018efficient} to generate fine-grained features of small UAVs could significantly improve detection performance, but comes at the cost of increased computational complexity.

    \item \textbf{Target confusion and occlusion in multi-UAV scenarios.} To address challenges related to occlusion and target similarity in multi-UAV tracking, future approaches may consider incorporating global detection modules  \cite{wang2018detect} to strengthen occlusion recovery. Furthermore, embedding self-attention mechanisms can improve the extraction of discriminative features, enabling the model to better distinguish between targets with similar appearances and motion patterns.
   However, such mechanisms often increase model complexity and may affect real-time performance in embedded systems, requiring further trade-off analysis.
    
    \item \textbf{Model robustness and false detection suppression.} In the context of critical infrastructure protection, enhancing the robustness of detection models and reducing false detection rates are essential for ensuring system reliability. Future methods could explore the design of dynamic loss functions (e.g., \cite{wu2018learning}, \cite{wang2022comprehensive}) that adaptively re-weight learning signals, particularly for small target detection, thereby improving the model's accuracy in distinguishing true targets from noise.

    \item \textbf{Modeling Structural Deformation and Dynamic Appearance for UAV.} Emerging foldable and morphable UAVs present new challenges for traditional visual tracking due to their changing structures during flight (e.g., folding wings, retractable parts) \cite{wang2025dist}. Current methods lack the ability to systematically handle such deformation. Future work should explore multi-form graph neural networks or deformation-aware models to enable continuous recognition under varying UAV shapes, which is essential for countering next-generation camouflaged and shape-shifting UAV threats.

    \item \textbf{Legal, Ethical, and Privacy Considerations.} As anti-UAV systems enter civilian and sensitive domains, it is crucial to ensure compliance with airspace laws, data privacy regulations, and operational rules. Technologies such as RF jamming and visual tracking may raise legal and ethical concerns, especially in populated areas. Future systems must embed legal constraints into hardware and algorithms, with clear threat evaluation, data limitations, and controlled countermeasure authorization. Addressing these issues is vital for the responsible and lawful deployment of anti-UAV technologies.

\end{itemize}

\section{Conclusion}
In this review, we comprehensively explore the latest advancements, challenges, and future research directions in Anti-UAV detection and tracking technology.  We first analyze the main features of current Anti-UAV detection and tracking technologies, including computer vision-based methods and the application of fusion detection and tracking algorithms. Through summarizing and analyzing existing datasets, we provide researchers with links to several publicly available datasets to support their efficient development in addressing the challenges of Anti-UAV technology. Additionally, this article summarizes the major Anti-UAV detection and tracking algorithms proposed in recent years,  discussing their limitations in complex environments, such as target confusion, occlusion, and insufficient detection capabilities for small targets.

In analyzing the existing technologies, we point out the challenges faced by current Anti-UAV technologies, including balancing real-time performance and accuracy, multimodal data fusion and heterogeneous sensor collaboration, insufficient generalization ability in diverse environments, and target confusion and occlusion in multi-UAV scenarios. To address these challenges, this article proposes seven future research directions: balancing real-time performance and accuracy, multimodal data fusion and heterogeneous sensor collaboration, insufficient generalization in diverse environments, target confusion and occlusion in multi-UAV scenarios,  model robustness and false detection suppression, modeling structural deformation and dynamic appearance
for UAV and legal, ethical, and privacy considerations.

In summary, As UAV threats become increasingly complex, future research will continue to drive the development of this field, improving the overall performance and reliability of Anti-UAV systems and providing more effective protection for public and national security.  At the same time, we hope this article will be a helpful resource for readers in their research efforts.

\section{Ackonwledgements}
Thanks to the support by the Guangxi Natural Science Foundation (Grant No. 2024GXNSFAA010484), and the National Natural Science Foundation of China (No. 62466013), this work has been made possible.

\bibliographystyle{IEEEtran}
\bibliography{newRef}


\vspace{20pt}
\begin{IEEEbiography}[{\includegraphics[width=1in,height=1.25in,clip,keepaspectratio]{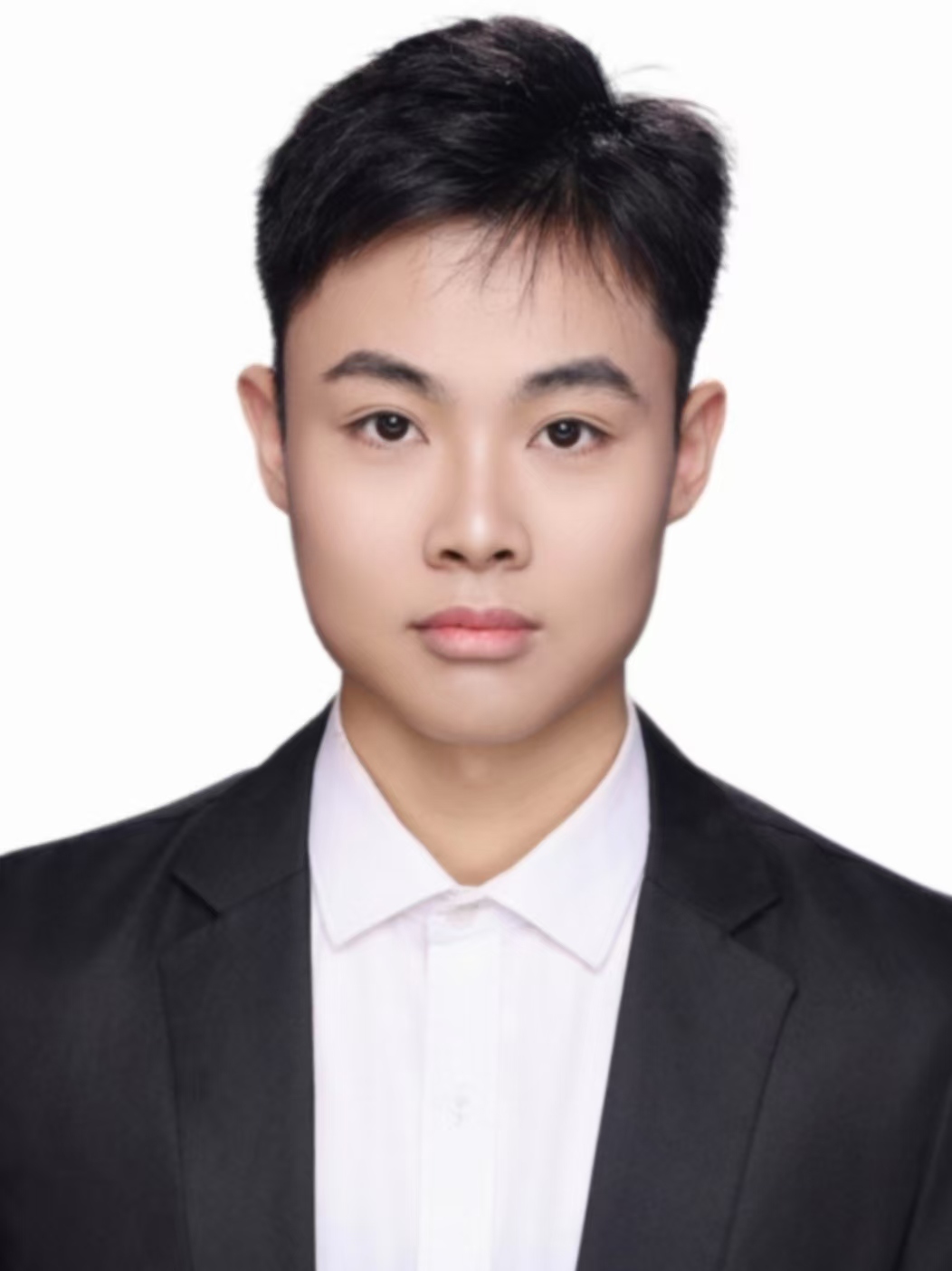}}]{Guanghai Ding}
received her B.Sc. degree in Software Engineering from Zhengzhou University of Science and Technology in 2024. He is currently a Master's degree candidate in the College of Computer Science and Engineering, Guilin University of Technology, Guilin, China. Her research interests are in computer vision, especially in object tracking.
\end{IEEEbiography}

\vspace{20pt}
\begin{IEEEbiography}[{\includegraphics[width=1in,height=1.25in,clip,keepaspectratio]{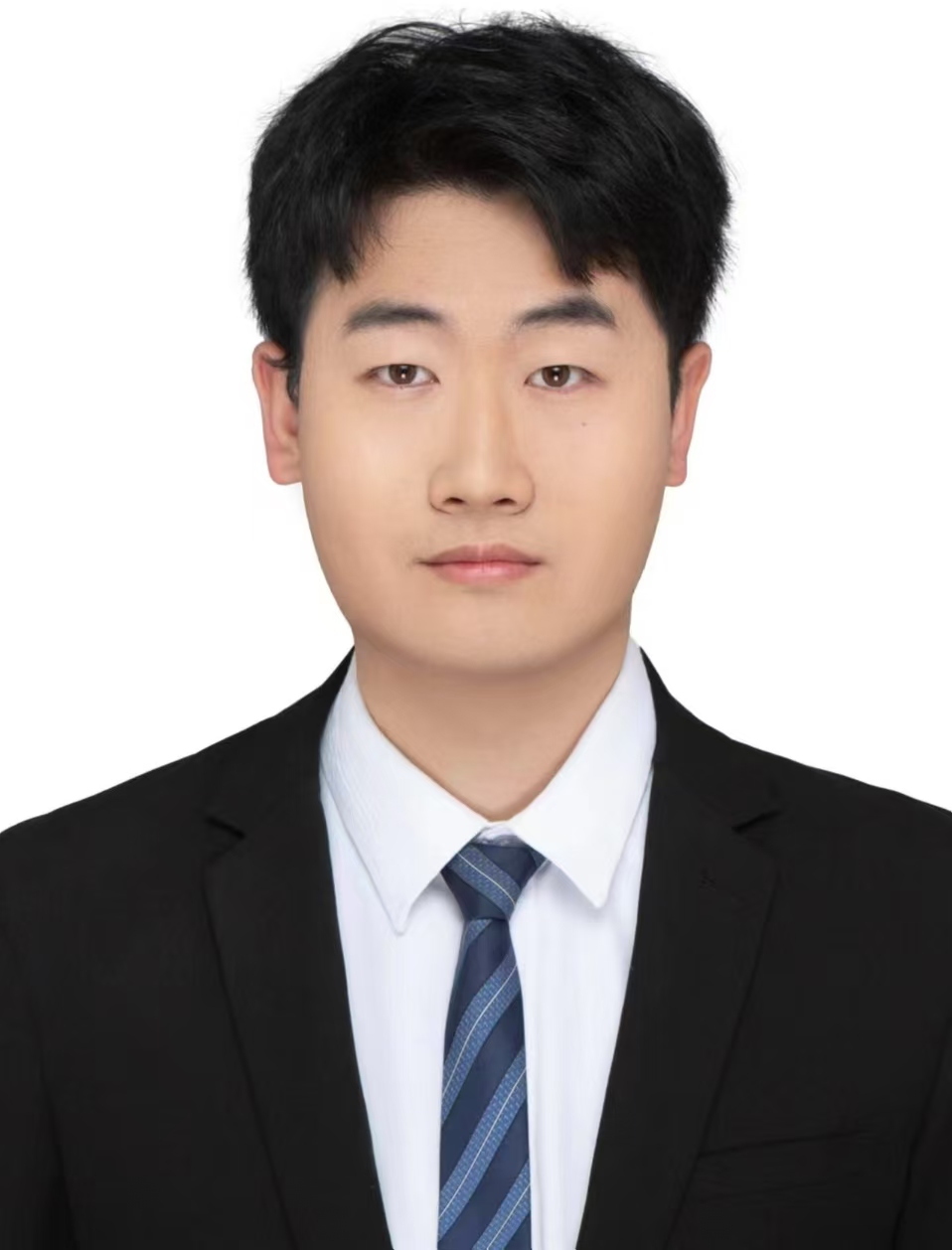}}]{Yihua Ren}
received the B.E. degree in computer science and technology from the School of Com puter Science, Northwestern Polytechnical Univer sity, Xi’an, China, in 2023. He is currently pursuing the M.E. degree with the School of Artificial In telligence, Optics and Electronics (iOPEN), North western Polytechnical University, Xi’an, China. His research interests include Anti-UAV tracking and transfer learning.
\end{IEEEbiography}

\vspace{20pt}
\begin{IEEEbiography}[{\includegraphics[width=1in,height=1.25in,clip,keepaspectratio]{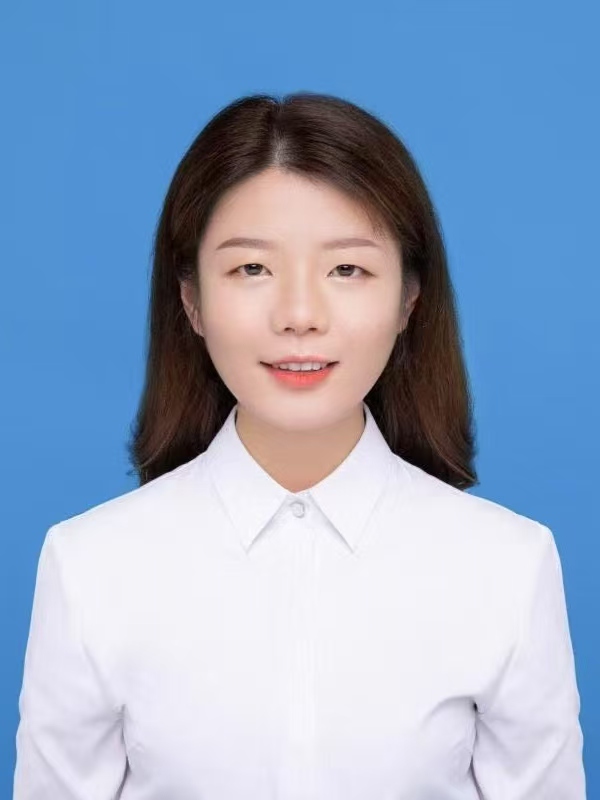}}]{Yuting Liu}
received the B.S. degrees from Nanchang University, China, in 2015. She has been in the Masters and PhD Combined Programs of Sichuan University. Her main research area is computer vision, specifically for challenges in crowd analysis.
\end{IEEEbiography}

\vspace{20pt}
\begin{IEEEbiography}[{\includegraphics[width=1in,height=1.25in,clip,keepaspectratio]{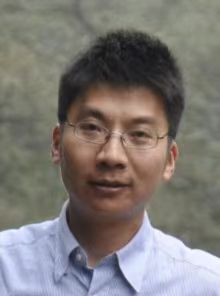}}]{Qijun Zhao} received the B.Sc. and M.Sc. Degrees in computer science from Shanghai Jiao Tong University, Shanghai, China, in 2003 and 2006, respectively, and the Ph.D. degree in computer science from The Hong Kong Polytechnic University, Hong Kong, in 2010. He was a Post-Doctoral Research Fellow with the Pattern Recognition and Image Processing Laboratory, Michigan State University, East Lansing, MI, USA, from 2010 to 2012. He is currently
a Professor with the College of Computer Science, Sichuan University, Chengdu, China. He is also a Visiting Professor with the School of Information Science and Technology, Tibet University, Tibet, China. His research is in the fields of pattern recognition, image processing, and computer vision.
\end{IEEEbiography}

\vspace{20pt}
\begin{IEEEbiography}[{\includegraphics[width=1in,height=1.25in,clip,keepaspectratio]{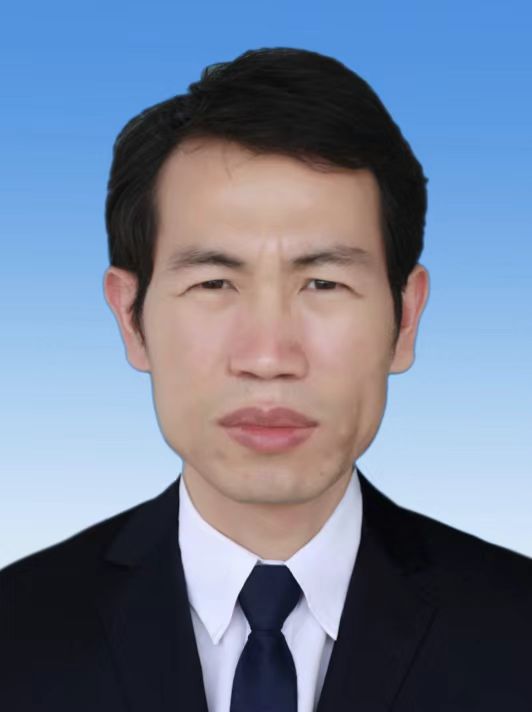}}]{Shuiwang Li}
received his Ph.D. degree in computer science and technology from Sichuan University in 2021. In 2015, he worked as a research assistant in the Institute of Computational and Theoretical Studies at Hong Kong Baptist University. He is currently an assistant professor in the College of Computer Science and Engineering, Guilin University of Technology, China. His research interests include pattern recognition, computer vision and machine learning.
\end{IEEEbiography}

\end{document}